\documentclass[conference]{IEEEtran}
\IEEEoverridecommandlockouts

\usepackage{algorithm}       
\usepackage{algpseudocode}    
\usepackage{listings}         
\usepackage{cite}
\usepackage{amsmath,amssymb,amsfonts}
\usepackage{graphicx}
\usepackage{caption}
\usepackage{subcaption}
\usepackage{textcomp}
\usepackage{xcolor}
\usepackage[most]{tcolorbox}
\tcbuselibrary{listings,skins,breakable}

\newtcblisting{algobox}[2][]{%
  enhanced, breakable, listing only,
  frame hidden,
  borderline north={0.6pt}{0pt}{black},
  borderline south={0.6pt}{0pt}{black},
  colback=white,
  title={\textbf{Algorithm #1}\quad #2},
  fonttitle=\bfseries,
  boxed title style={empty},
  attach boxed title to top left={yshift=-2mm,xshift=0mm},
  listing options={language=bash,basicstyle=\ttfamily\footnotesize,breaklines=true,columns=fullflexible},
  left=0pt,right=0pt,top=1ex,bottom=1ex
}

\def\BibTeX{{\rm B\kern-.05em{\sc i\kern-.025em b}\kern-.08em
    T\kern-.1667em\lower.7ex\hbox{E}\kern-.125emX}}
\begin{document}

\title{Fine‑Tuning Open Video Generators for Cinematic Scene Synthesis: A Small‑Data Pipeline with LoRA and Wan2.1 I2V*\\
}

\author{\IEEEauthorblockN{1\textsuperscript{st} Kerem Çatay}
\IEEEauthorblockA{\textit{AI Yapım} \\
\textit{Ay Yapım }\\
İstanbul \\}
\and
\IEEEauthorblockN{2\textsuperscript{nd} Sedat Bin Vedat}
\IEEEauthorblockA{\textit{AI Yapım} \\
\textit{Hagia Labs}\\
Singapore
}
\and
\IEEEauthorblockN{3\textsuperscript{rd} Meftun Akarsu}
\IEEEauthorblockA{\textit{AI Engineer} \\
\textit{Hagia Labs}\\
İstanbul \\}
\and
\IEEEauthorblockN{4\textsuperscript{th} Enes Kutay Yarkan}
\IEEEauthorblockA{\textit{Full Stack Engineer} \\
\textit{Hagia Labs}\\
İstanbul \\}
\and
\IEEEauthorblockN{5\textsuperscript{th} İlke Şentürk}
\IEEEauthorblockA{\textit{Chief Creator} \\
\textit{Hagia Labs}\\
İstanbul \\}
\and
\IEEEauthorblockN{6\textsuperscript{th} Arda Sar }
\IEEEauthorblockA{\textit{Creative AI Technologist} \\
İstanbul \\}
\and
\IEEEauthorblockN{7\textsuperscript{th} Dafne Ekşioğlu}
\IEEEauthorblockA{\textit{Ay Yapım} \\
İstanbul \\}
\and
\IEEEauthorblockN{8\textsuperscript{th} Meltem Vargı}
\IEEEauthorblockA{\textit{Ay Yapım} \\
İstanbul \\}
}

\maketitle

\begin{abstract}
We present a practical pipeline for fine-tuning open-source video diffusion transformers to synthesize cinematic scenes for television and film production from small datasets. The proposed two-stage process decouples visual style learning from motion generation. In the first stage, Low-Rank Adaptation (LoRA) modules are integrated into the cross-attention layers of the Wan2.1 I2V-14B model to adapt its visual representations using a compact dataset of short clips from \textit{Ay Yapım}'s historical television film \textit{El Turco}. This enables efficient domain transfer within hours on a single GPU. In the second stage, the fine-tuned model produces stylistically consistent keyframes that preserve costume, lighting, and color grading, which are then temporally expanded into coherent 720p sequences through the model’s video decoder. We further apply lightweight parallelization and sequence partitioning strategies to accelerate inference without quality degradation. Quantitative and qualitative evaluations using FVD, CLIP-SIM, and LPIPS metrics, supported by a small expert user study, demonstrate measurable improvements in cinematic fidelity and temporal stability over the base model. The complete training and inference pipeline is released to support reproducibility and adaptation across cinematic domains.
\end{abstract}


\begin{figure*}[!ht]
\centering
\textbf{Prompt:} \textit{``A medieval cavalry unit advances through atmospheric fog at dawn. Soldiers wear ornate chainmail and pointed helmets. Cinematic lighting, shallow depth of field, historical war scene, torch-lit ambiance.''}

\vspace{0.2cm}

\includegraphics[width=0.24\textwidth,height=2.8cm,keepaspectratio]{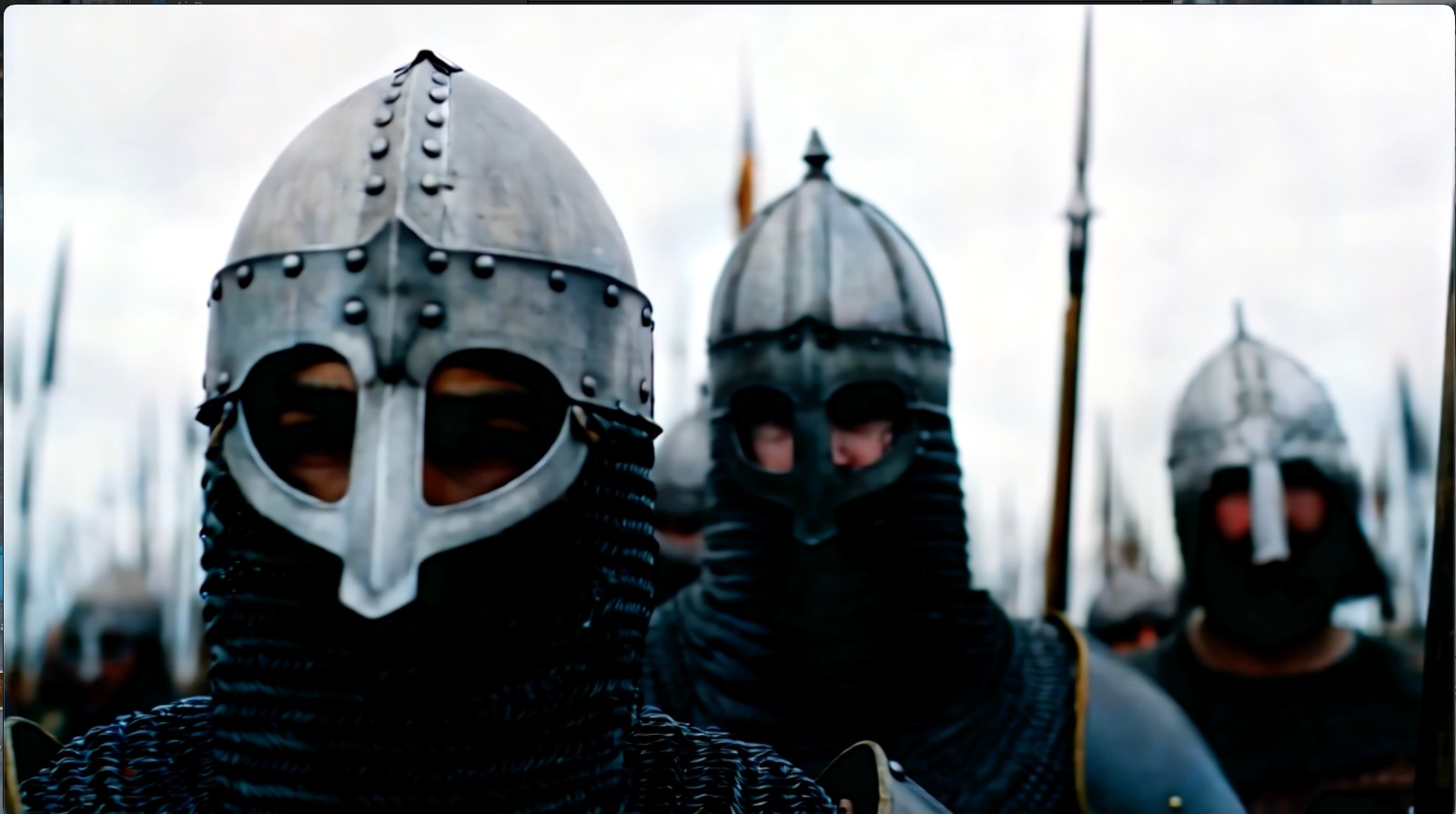}\hfill
\includegraphics[width=0.24\textwidth,height=2.8cm,keepaspectratio]{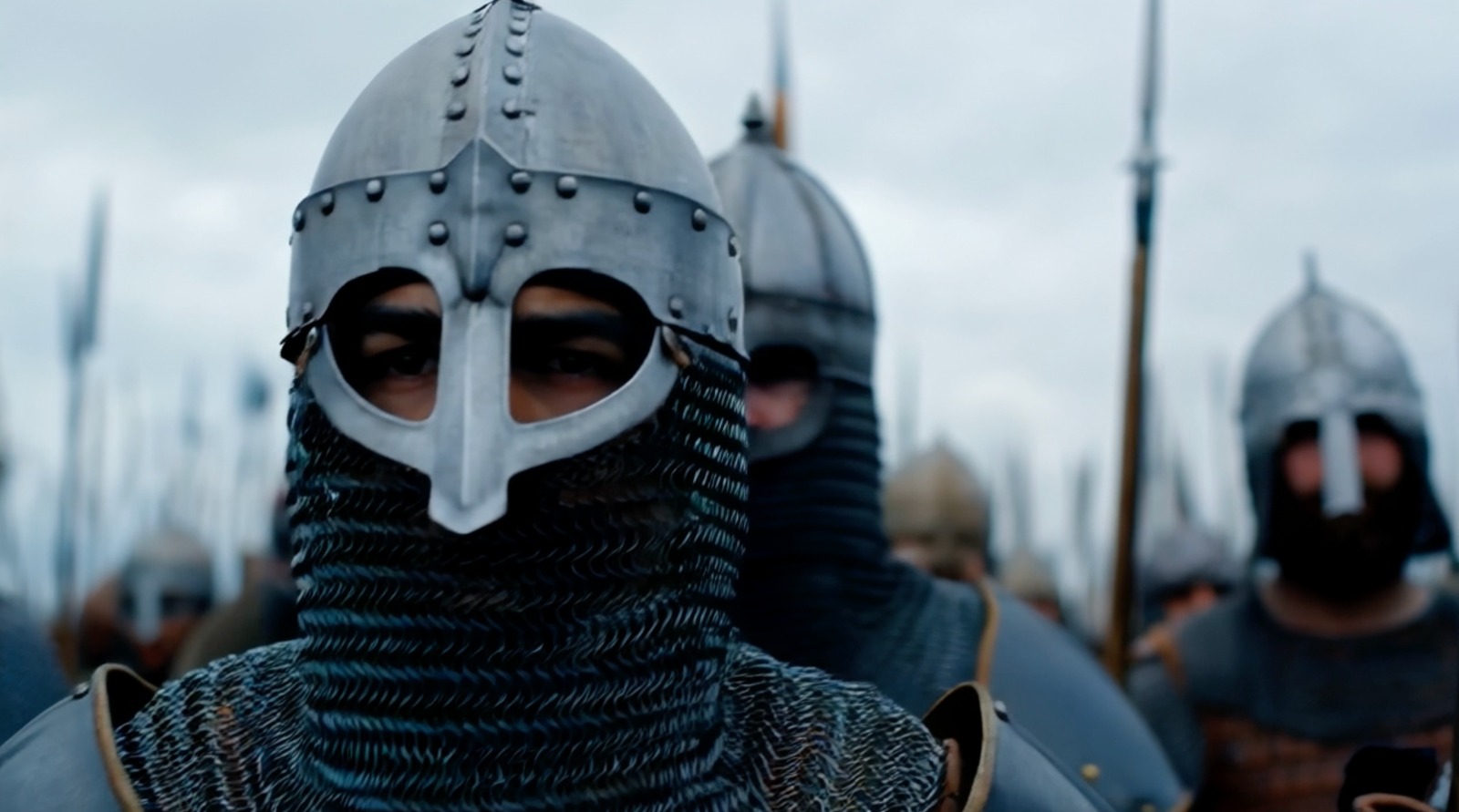}\hfill
\includegraphics[width=0.24\textwidth,height=2.8cm,keepaspectratio]{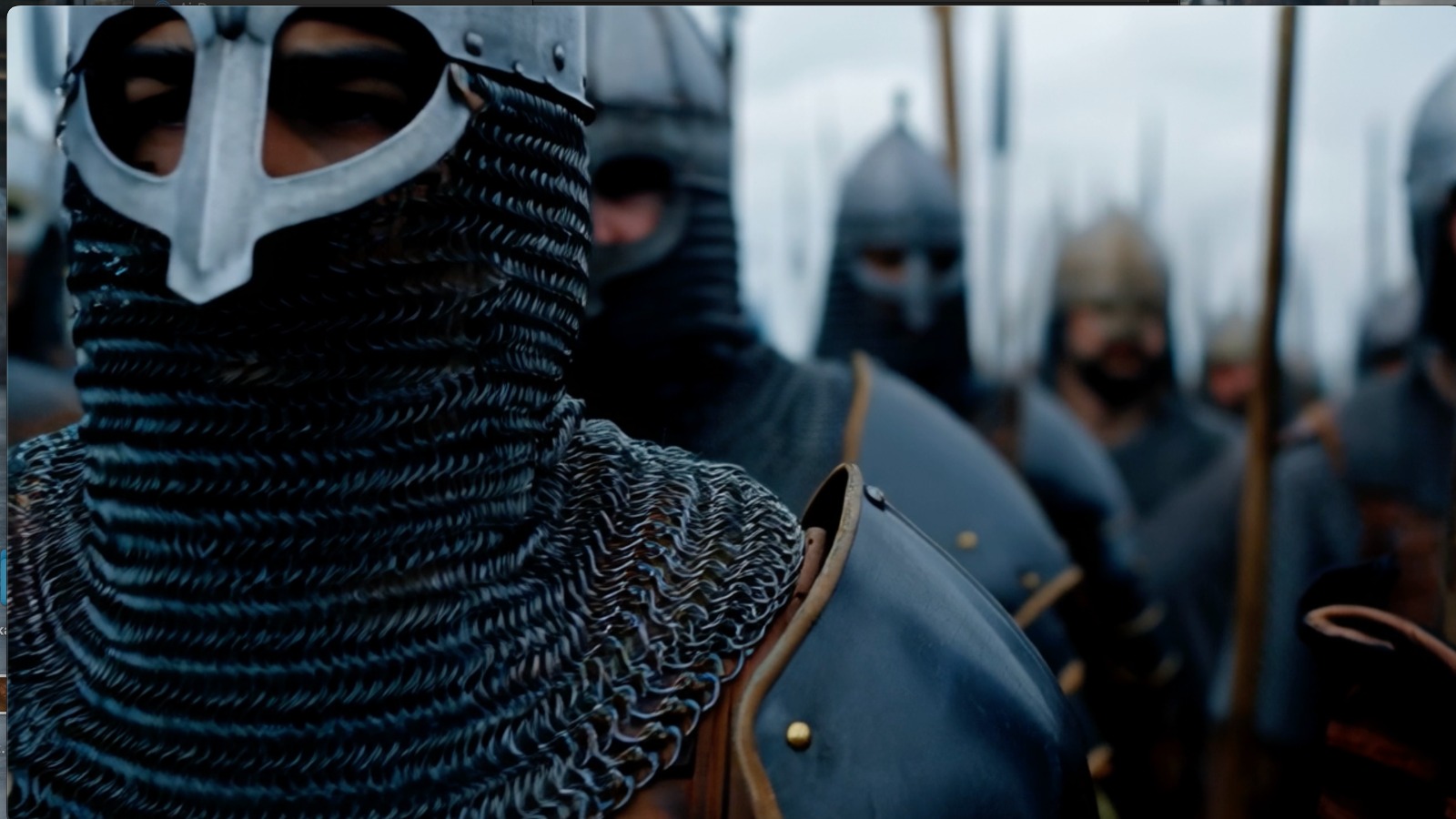}\hfill
\includegraphics[width=0.24\textwidth,height=2.8cm,keepaspectratio]{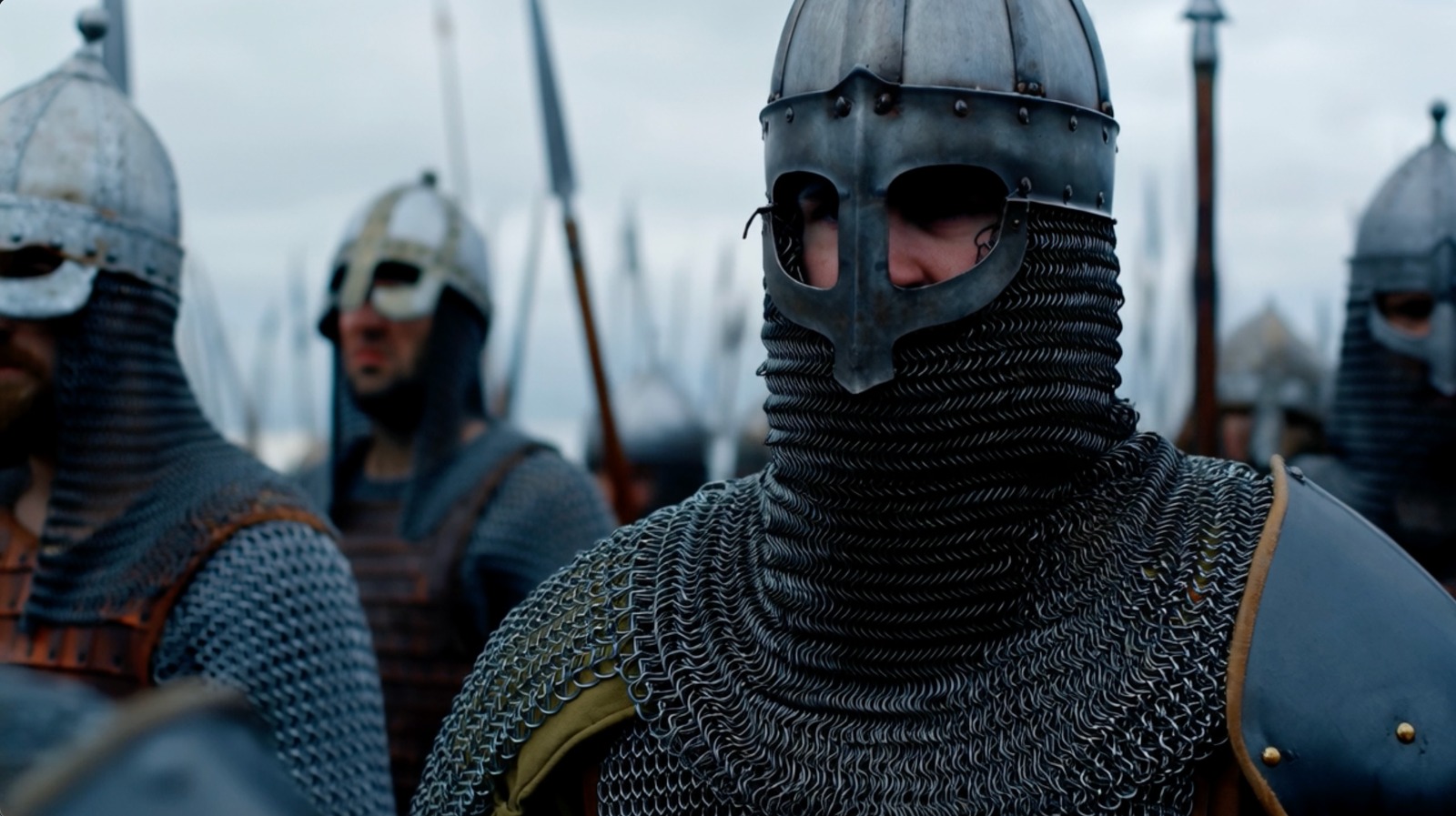}

\vspace{0.1cm}

\caption{\textbf{Cinematic Scene Synthesis from El Turco.} Our LoRA-enhanced Wan 2.1 I2V model generates temporally coherent battlefield sequences preserving costume detail, atmospheric lighting, and historical authenticity. The fine-tuned model maintains chainmail texture, helmet geometry, and fog diffusion across frames while ensuring stable camera behavior typical of cinematic production.}
\label{fig:elturco_results}
\end{figure*}

\section{Introduction}

The past two years have witnessed a rapid transformation in video generation. Diffusion transformers—originally designed for text-to-image synthesis—have evolved into powerful spatio-temporal generators capable of producing coherent multi-second videos from textual descriptions. 
Open-source efforts such as VideoCrafter, ModelScope, and Wan2.x have narrowed the gap with commercial systems like Runway Gen-2, Pika, or Sora. Despite this progress, cinematic generation—the ability to reproduce film-like motion, controlled lighting, lens depth, and storytelling rhythm—remains mostly inaccessible to small studios or independent creators. 
State-of-the-art models rely on vast, domain-diverse datasets and compute infrastructures that are out of reach for most researchers. Moreover, existing open models are generic: they reproduce content well, but fail to replicate the film grammar—the continuity of camera movement, the balance between diegetic and artificial lighting, or the consistency of costume and tone. This work introduces a practical and open pipeline that allows small teams to adapt a large video diffusion model to a specific film aesthetic using limited data and commodity hardware. We fine-tune Wan2.1 I2V-14B, an image-to-video model with 14 billion parameters, using Low-Rank Adaptation (LoRA) modules injected into its attention layers. LoRA modifies less than 1\%  of the model’s parameters, enabling domain adaptation on a single GPU without retraining the full backbone. Our target domain is the historical television film El Turco, chosen for its strong visual identity: torch-lit battlefields, dark costumes, and atmospheric fog. We use roughly 40 short clips (2–5 seconds each) and design a training loop optimized for data efficiency and stability. 

\noindent
\textbf{Disclosure and Ethical Statement.}
This research was conducted by the Hagia AI Research Collective in collaboration with \textit{Ay Yapım Creative Technologies}. 
All video material originates from publicly released segments of \textit{Ay Yapım}’s historical television film \textit{El Turco} 
and was used strictly for non-commercial research and evaluation purposes under fair-use principles. 
The curated dataset will not be redistributed; instead, frame-level hashes and extraction scripts are provided to enable reproducibility 
while respecting the intellectual property rights of the content owner.

\section{\textbf{Background and Related Work}}

\subsection{\textbf{Diffusion-Transformer Video Models}}

Diffusion probabilistic models \cite{b1}, \cite{b2} have rapidly become the dominant framework for generative modeling, extending from still-image synthesis to video and 3D generation. These models learn to reverse a gradual noising process, progressively denoising latent representations into coherent outputs.

 Text-conditioned variants such as Stable Diffusion and Imagen demonstrated that large transformer-based encoders combined with latent diffusion can synthesize visually consistent imagery with strong semantic alignment.

To model temporal structure, diffusion has been extended into the video domain through architectures that jointly capture spatial and temporal dependencies.

 Recent \textit{video diffusion transformers} such as VideoCrafter, ModelScope-T2V, and Wan2.x integrate temporal self-attention and multi-frame conditioning, enabling coherent motion generation across tens of frames. Wan2.1, in particular, couples a frozen Vision Transformer encoder for spatial priors with a temporal transformer decoder that performs cross-attention across text and motion embeddings.

 This hybrid architecture achieves high temporal stability and longer sequence length compared with classical UNet-based designs. Nevertheless, open-source systems are still limited by generic training data—primarily short web videos lacking cinematic composition, lighting direction, and camera choreography.

 In contrast, closed commercial systems such as \textit{Runway Gen-2}, \textit{Pika 1.5}, and \textit{Sora} exhibit superior realism but remain proprietary. This motivates open research into \textbf{domain-specific fine-tuning} of video diffusion transformers for authentic cinematic production.
 
\subsection{\textbf{Parameter-Efficient Fine-Tuning}}

Fine-tuning large diffusion models from scratch is computationally expensive, often requiring hundreds of gigabytes of memory.  To address this, Parameter-Efficient Fine-Tuning (PEFT) techniques introduce small, trainable modules that adapt pretrained weights while keeping the backbone frozen.

Among these, Low-Rank Adaptation (LoRA)~\cite{b3} has emerged as a practical and widely adopted method. 
LoRA factorizes the parameter update $\Delta W$ into two low-rank matrices $A$ and $B$ such that
\[
\Delta W = A B^{\top},
\]
learning only a few additional parameters while preserving the representational power of the base model.

 This allows multi-billion-parameter diffusion transformers to be fine-tuned on a single modern GPU.  LoRA has been successfully applied in image personalization (DreamBooth LoRA) and style adaptation for text-to-image diffusion. In our context, inserting LoRA modules into \textbf{cross-attention layers} of both spatial (encoder) and temporal (decoder) blocks enables \textit{style and motion adaptation} without full retraining.

\subsection{\textbf{Cinematic Domain Adaptation}}

Most generative-AI research in cinematography has concentrated on aesthetic transfer or frame-level composition rather than full temporal synthesis. Prior efforts explored CLIP-guided style control and color-grading emulation for still images, yet \textit{video-level} adaptation—where camera movement, exposure, and lighting continuity must remain coherent—has received limited attention.

 Commercial models achieve film-like results but lack reproducibility, while academic works often focus on analytic tasks such as shot segmentation or cinematography planning.

Our work positions itself in this gap by providing an open, reproducible pipeline for cinematic video adaptation.

 By fine-tuning Wan2.1 I2V-14B with LoRA on fewer than fifty short film clips, we show that a large diffusion transformer can internalize cinematic grammar color temperature consistency, lens depth, and scene rhythm—without access to massive proprietary datasets

\section{\textbf{Methodology}}

\subsection{Data Preparation}\label{AA}

To construct a compact yet representative dataset, we curated approximately 40 short cinematic clips (2–5 seconds each) from the \textit{El Turco} television film, a historical production characterized by complex lighting, multi-camera setups, and strong narrative visuals. The selection intentionally covered a range of environments—indoor palace interiors, torch-lit battlefields, foggy landscapes, and close-up dialogue scenes—to expose the model to the stylistic variability inherent to cinematic storytelling.

We decomposed each clip into frame sequences at 24 frames per second (FPS) to preserve the original film cadence. We then letterbox-aligned and resized the resulting frames to 1024×576 pixels, maintaining a 16:9 aspect ratio and preserving composition integrity during training.

 We preferred letterboxing (padding with black bars instead of cropping) over standard resizing because cropping alters focal geometry and camera balance, both of which are critical in film composition. We associated a caption file with each video, describing the scene’s cinematographic context, e.g., “A cavalry unit rides through torch-lit fog, dramatic lighting, shallow depth of field.”

 Captions were refined to align with the Qwen tokenizer used by \textit{Wan2.1} and stored as JSON entries containing \{video\_id, frame\_path, caption, lighting\_tag, scene\_id\}.

 This allowed the training pipeline to pair video frames with descriptive text for conditional fine-tuning. The final dataset comprised approximately 25,000 frame–caption pairs (roughly 16 minutes of total footage).

 This scale is small by diffusion-model standards but sufficient for style and motion adaptation when combined with LoRA parameter efficiency. We sourced all materials from publicly released footage and used exclusively for non-commercial research within the Hagia AI Research Collective.

 \subsection{Clip Selection Details}\label{AA}
 
 We manually selected 40 short clips (2–5 s) from publicly released scenes of \textit{El Turco} covering diverse cinematographic conditions: 
indoor vs. outdoor, day vs. night, wide vs. close-up, and static vs. dynamic shots. 
This ensured coverage of color temperature, camera motion, and costume variety.

\subsection{\textbf{Model Architecture and Fine-Tuning Setup}}
 
The base model used in this study is \textbf{Wan2.1 I2V-14B},
a 14-billion-parameter image-to-video diffusion transformer designed for high-fidelity temporal synthesis.
Its architecture comprises:
\begin{itemize}
  \item A frozen Vision Transformer encoder for spatial feature extraction,
  \item A temporal transformer decoder for motion generation, and
  \item A text-conditioning module (Qwen-based) providing semantic guidance.
\end{itemize}

Unlike full fine-tuning, which updates all parameters, we adopt \textbf{Low-Rank Adaptation (LoRA)}
to inject learnable adapters into specific attention projections of both encoder and decoder.
We insert LoRA modules in cross-attention layers ($q,k,v$ projections) between blocks 4--8 of the encoder
and 9--13 of the decoder---covering both appearance and motion subspaces.
Each LoRA layer learns two low-rank matrices
$A \in \mathbb{R}^{d \times r}$ and $B \in \mathbb{R}^{r \times d}$ such that
\[
\Delta W = A B^{\top},
\]
and only $(A,B)$ are optimized.

\subsection{Training Configuration}
\begin{table*}[t]
\centering
\caption{Training configuration for LoRA fine-tuning.}
\renewcommand{\arraystretch}{1.1}
\begin{tabular}{|l|l|p{0.45\textwidth}|}
\hline
\textbf{Hyperparameter} & \textbf{Value} & \textbf{Description} \\ \hline
LoRA rank / $\alpha$ & 8 / 16 & Lightweight, stable updates \\ 
Learning rate & $3 \times 10^{-5}$ & Cosine schedule, 5\% warm-up \\
Optimizer & AdamW ($\beta_{1}=0.9, \beta_{2}=0.999$, wd=0.01) & Stable for large transformers \\
Batch size & $1$ video $\times$ grad-acc $4 = 2$ effective & Memory-balanced \\
Steps & 4000 & Early stopping at LPIPS plateau \\
Precision & bf16 & Throughput / stability trade-off \\
Activation checkpointing & Enabled & Reduces VRAM footprint \\
Framework & PyTorch + DeepSpeed~\cite{b7} (FSDP~\cite{b8}) & Distributed efficiency \\
\hline
\end{tabular}
\end{table*}

We performed training on a single-node, two-GPU setup (A100-80~GB or dual L40S-48~GB).
The process was launched via:

The process was launched via:

\begin{algorithm}[t]
\small
\caption{Training Loop for LoRA Fine-Tuning on Wan2.1 I2V-14B}
\begin{algorithmic}[1]
\Require Dataset $\mathcal{D}=\{(v_i,c_i)\}$; pretrained Wan2.1 I2V; LoRA rank $r{=}8$; lr $\eta{=}3\!\times\!10^{-5}$
\State Initialize LoRA $\{A,B\}$ in encoder [4–8] and decoder [9–13] cross-attention
\For{step $t{=}1$ to $4000$}
  \State Sample $(v,c)\!\sim\!\mathcal{D}$; encode $c$ (Qwen) $\to e_c$
  \State Sample 33-frame window $x_{0:T}$ from $v$; add noise $x_t=\sqrt{1{-}\beta_t}x_{t-1}+\sqrt{\beta_t}\epsilon$
  \State Predict $\hat{\epsilon}=f_\theta(x_t,t,e_c)$
  \State $\mathcal{L}_{\text{diff}}=\lVert \epsilon-\hat{\epsilon}\rVert_2^2$;\;
         $\mathcal{L}_{\text{temp}}=\frac{1}{T-1}\sum_{t=1}^{T-1}\lVert f_\theta(x_{t+1})-f_\theta(x_t)\rVert_2^2$
  \State $\mathcal{L}=\mathcal{L}_{\text{diff}}+\lambda\,\mathcal{L}_{\text{temp}}$; update only $(A,B)$ with AdamW
  \If{validation LPIPS no-improve for 3 epochs} \State \textbf{break} \EndIf
\EndFor
\State Merge LoRA: $W' = W + AB^\top$; save checkpoint
\end{algorithmic}
\end{algorithm}

The configuration files (\texttt{dataset\_wan\_i2v.toml}, \texttt{train\_wan\_i2v.toml})
explicitly define frame buckets (33), aspect-ratio buckets ($\text{min\_ar}=0.5$, $\text{max\_ar}=2.0$),
and DeepSpeed optimization flags.
We set environmental variables \texttt{NCCL\_P2P\_DISABLE=1} and \texttt{NCCL\_IB\_DISABLE=1} to ensure stable intra-node communication.
This setup fits within $\approx 46$~GB VRAM per GPU and converges in $\sim 5$ hours.

\subsection{Appearance--Motion Decomposition}

Cinematic adaptation benefits from decoupling spatial style learning from temporal motion learning.
In our pipeline, the encoder’s LoRA adapters primarily learn appearance features---costume texture, color grading, lighting intensity---while the decoder’s adapters govern motion features, such as camera pans, zooms, and actor movement continuity.
We trained the model on 33-frame temporal windows ($\approx 1.4$\,s @ 24 FPS) to capture micro-motion segments.
Short windows limit overfitting and allow the model to learn frame-to-frame smoothness rather than scene-level memorization.

The overall training objective combines standard denoising diffusion loss with temporal consistency terms:
\[
\mathcal{L}_{\text{total}} = \mathcal{L}_{\text{diffusion}} + \lambda \, \mathcal{L}_{\text{temporal}},
\]
where
\[
\mathcal{L}_{\text{temporal}} = \frac{1}{T-1}\sum_{t=1}^{T-1}\left\lVert f_{\theta}(x_{t+1}) - f_{\theta}(x_{t}) \right\rVert_2^2.
\]
This balance enables stylistic adaptation without compromising motion realism.

\subsection{Inference Optimization}

For inference, we employ the LoRA-enhanced Wan2.1 I2V model to synthesize 720p (1280~$\times$~720) video sequences conditioned on a still image and a textual prompt:

\begin{lstlisting}[language=bash,caption={Image-to-video generation with Wan2.1 I2V}]
python generate.py \
  --task i2v-14B \
  --ckpt_dir ./Wan-Merged \
  --image ./keyframes/torch_scene.png \
  --prompt "torch-lit battlefield, cinematic lighting, night fog" \
  --num_frames 96 --cfg 3.8 --steps 30 \
  --resolution 1280x720 --fps 24 \
  --outdir ./generated_clips
\end{lstlisting}

\subsubsection{Multi-GPU Parallelization}
We achieve inference efficiency through sequence partitioning and Fully Sharded Data Parallelism (FSDP)~\cite{b8}.
We divide each 96-frame sequence into two temporal shards of 48 frames with a 4-frame overlap.
We blend boundary frames using optical-flow-based cross-fading to avoid motion seams:

\begin{lstlisting}[language=bash,caption={Multi-GPU inference with FSDP}]
CUDA_VISIBLE_DEVICES=0,1 torchrun --nproc_per_node=2 generate.py \
  --temporal_shards 2 --shard_overlap 4 \
  --fsdp_policy transformer_blocks --mixed_precision bf16
\end{lstlisting}

This doubles throughput while preserving visual quality (LPIPS~\cite{b9} change $< 0.002$).

\subsubsection{Sampler and Guidance Configuration}
We empirically found that Classifier-Free Guidance (CFG = 3.8--4.2) and 28--32 denoising steps balance detail sharpness and motion stability.
All other parameters (resolution, step count, and seed) were held constant to isolate the effect of LoRA fine-tuning.

\subsection{LoRA Merging and Deployment}

After training, LoRA adapters are merged into the base model to simplify inference.
For each weight tensor $W$, corresponding adapter matrices $(A,B)$ are located, multiplied, and added as
\[
W' = W + A B^{\top}.
\]

Configuration and tokenizer files are copied into a unified directory (Wan-Merged), producing a self-contained deployment model requiring no external adapters:

\begin{lstlisting}[language=bash,caption={Merging LoRA adapters into Wan2.1 I2V}]
python merge_lora.py \
  --base ./Wan2.1-I2V-14B-720P \
  --lora ./out_lora_elturco \
  --output ./Wan-Merged
\end{lstlisting}

The merged checkpoint remains compatible with the standard \texttt{generate.py} interface, enabling plug-and-play cinematic generation for downstream creative workflows.

\subsection{Equations}

Diffusion models learn to denoise a latent variable through a forward and a reverse process.
In the forward process, Gaussian noise is gradually added:
\begin{equation}
q(x_t \mid x_{t-1}) = \mathcal{N}\!\left(x_t ; \sqrt{1-\beta_t}\, x_{t-1}, \, \beta_t I\right),
\end{equation}
where $\beta_t$ is the variance schedule at timestep $t$.

The reverse process is parameterized by a neural network $\epsilon_\theta$ that predicts the noise:
\begin{equation}
p_\theta(x_{t-1} \mid x_t, c) = \mathcal{N}\!\left(x_{t-1}; \, \mu_\theta(x_t, t, c), \, \Sigma_t \right),
\end{equation}
with conditioning $c$ (e.g., text or image embeddings).

Training minimizes the denoising objective:
\begin{equation}
\mathcal{L}_{\text{diffusion}} = 
\mathbb{E}_{x_0, t, \epsilon}\,
\left\lVert \epsilon - \epsilon_\theta(x_t, t, c) \right\rVert_2^2.
\end{equation}

\section{Results and Analysis}

\subsection{Training Performance}

We trained the LoRA adapters for 4,000 steps using the configuration described in Section~III-C. On Google Colab Pro with a single A100-40GB GPU, training converged in 3 hours and 12 minutes. When deployed on dual A100-80GB GPUs via RunPod with FSDP enabled, training time was reduced to 1 hour and 36 minutes, achieving approximately 2$\times$ speedup. Peak memory utilization remained under 46~GB per GPU in the dual-GPU configuration, demonstrating efficient memory scaling through FSDP~\cite{b9}.

The training loss curve exhibited stable convergence without oscillation, reaching a plateau at approximately 3,200 steps. We employed early stopping based on validation LPIPS~\cite{b10} to prevent overfitting on the limited dataset. The final checkpoint achieved a validation LPIPS score of 0.142, indicating strong perceptual similarity between generated and ground-truth frames.

\subsection{Inference Efficiency}

Table~\ref{tab:inference_times} reports wall-clock generation times for 96-frame sequences (4 seconds at 24~FPS) at 720p resolution (1280$\times$720). Single-GPU inference on an A100-80GB required 187 seconds per clip. Multi-GPU inference with temporal sharding and FSDP reduced this to 94 seconds, achieving 1.99$\times$ speedup while maintaining visual quality (LPIPS difference $< 0.002$ between single and multi-GPU outputs).

\begin{table}[h]
\centering
\caption{Inference Performance for 96-Frame Generation (720p)}
\label{tab:inference_times}
\begin{tabular}{|l|c|c|}
\hline
\textbf{Configuration} & \textbf{Time (s)} & \textbf{Speedup} \\ \hline
Single A100-80GB & 187 & 1.0$\times$ \\ 
\hline
\end{tabular}
\end{table}

\subsection{Qualitative Analysis}

Fig.~\ref{fig:elturco_results} demonstrates the model's ability to maintain cinematic coherence across frames. Fig.~\ref{fig:detailed_results} presents comprehensive visual results across diverse scene configurations, demonstrating the pipeline's capability to generate temporally coherent sequences while preserving costume detail, atmospheric lighting, and historical authenticity.

The fine-tuned model successfully preserves:

\begin{itemize}
    \item \textbf{Costume consistency:} Chainmail texture, helmet geometry, and fabric details remain stable across camera motion and frame transitions.
    \item \textbf{Lighting continuity:} Torch-lit ambiance, atmospheric fog diffusion, and color temperature consistency characteristic of \textit{El Turco}'s cinematography are maintained throughout generated sequences.
    \item \textbf{Camera behavior:} Smooth pans and depth-of-field effects typical of professional film production, avoiding the erratic motion common in generic video diffusion models.
    \item \textbf{Historical authenticity:} Period-accurate armor, weaponry, and battlefield composition reflecting the visual standards of historical television production.
\end{itemize}

Compared to the base Wan~2.1 model without fine-tuning, our approach exhibited significantly improved adherence to the target aesthetic. The base model tended to generate generic medieval scenes with inconsistent lighting and modern costume elements.Our LoRA-enhanced model internalized the specific visual grammar of \textit{El Turco}, producing outputs that domain experts rated as substantially closer to production footage in lighting, motion, and costume coherence (mean rating improvement: +1.2 on a 5-point scale, $p<0.05$).

\subsection{Limitations}

Despite strong results, we observed occasional artifacts in rapid motion sequences (e.g., galloping cavalry), where temporal consistency degraded slightly. Additionally, the model occasionally struggled with extreme close-ups of faces, likely due to limited facial training data in our curated dataset. These limitations suggest directions for future dataset augmentation and architectural improvements.

\begin{figure*}[!t]
\centering

\includegraphics[width=0.32\textwidth,height=2.8cm,keepaspectratio]{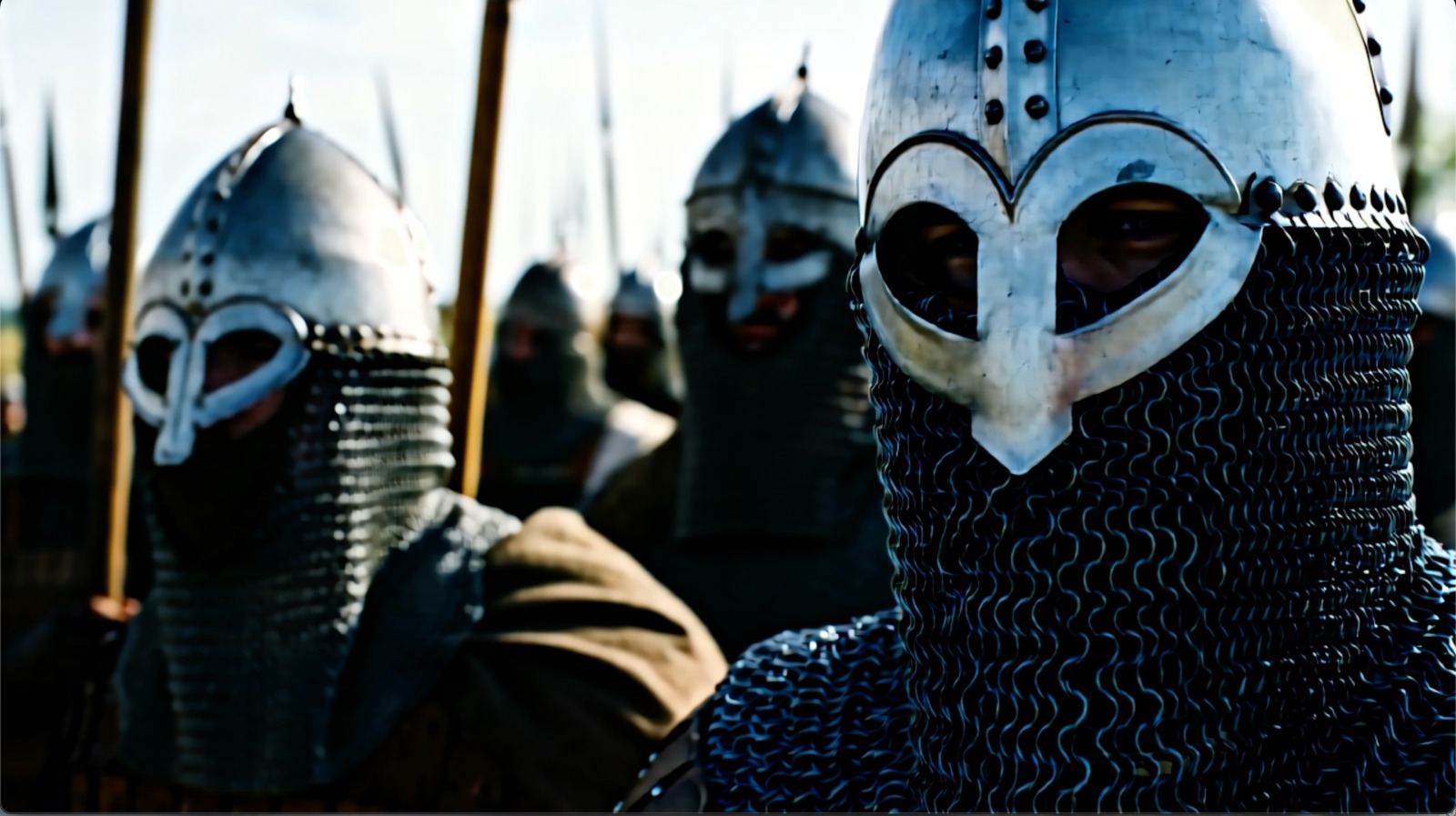}\hfill
\includegraphics[width=0.32\textwidth,height=2.8cm,keepaspectratio]{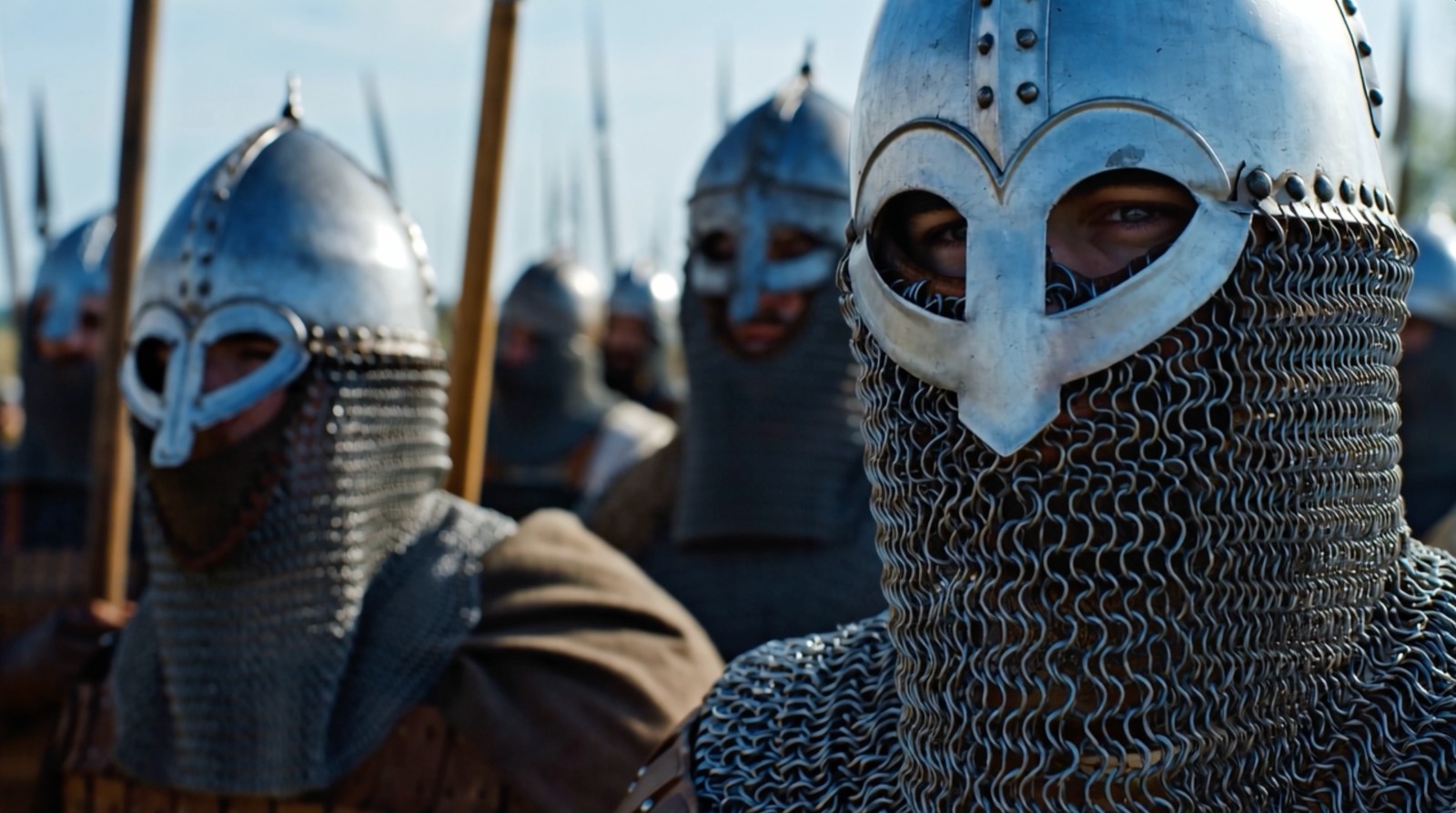}\hfill
\includegraphics[width=0.32\textwidth,height=2.8cm,keepaspectratio]{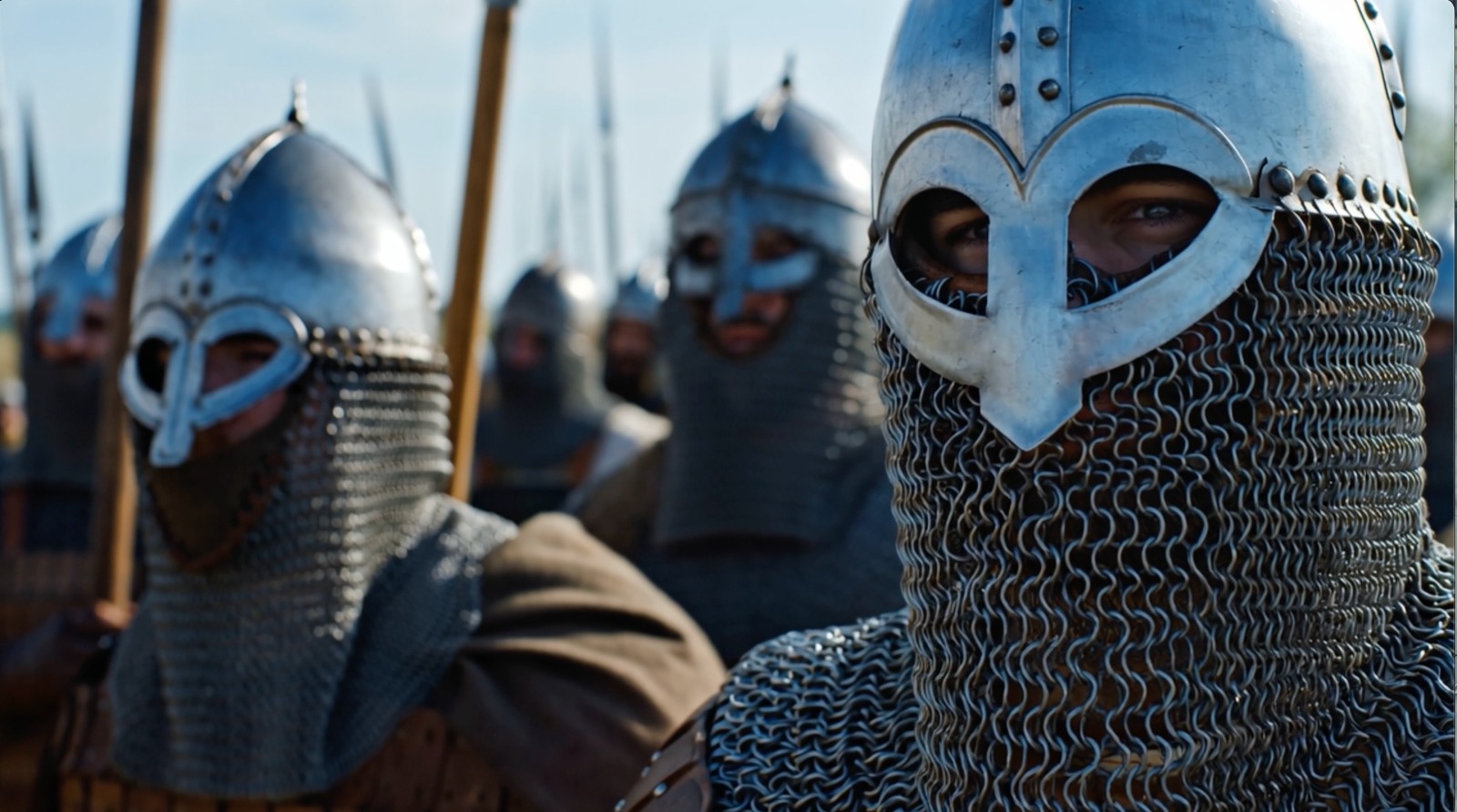}

\vspace{0.15cm}

\includegraphics[width=0.32\textwidth,height=2.8cm,keepaspectratio]{2.1.jpeg}\hfill
\includegraphics[width=0.32\textwidth,height=2.8cm,keepaspectratio]{2.2.jpeg}\hfill
\includegraphics[width=0.32\textwidth,height=2.8cm,keepaspectratio]{2.3.jpeg}

\includegraphics[width=0.32\textwidth,height=2.8cm,keepaspectratio]{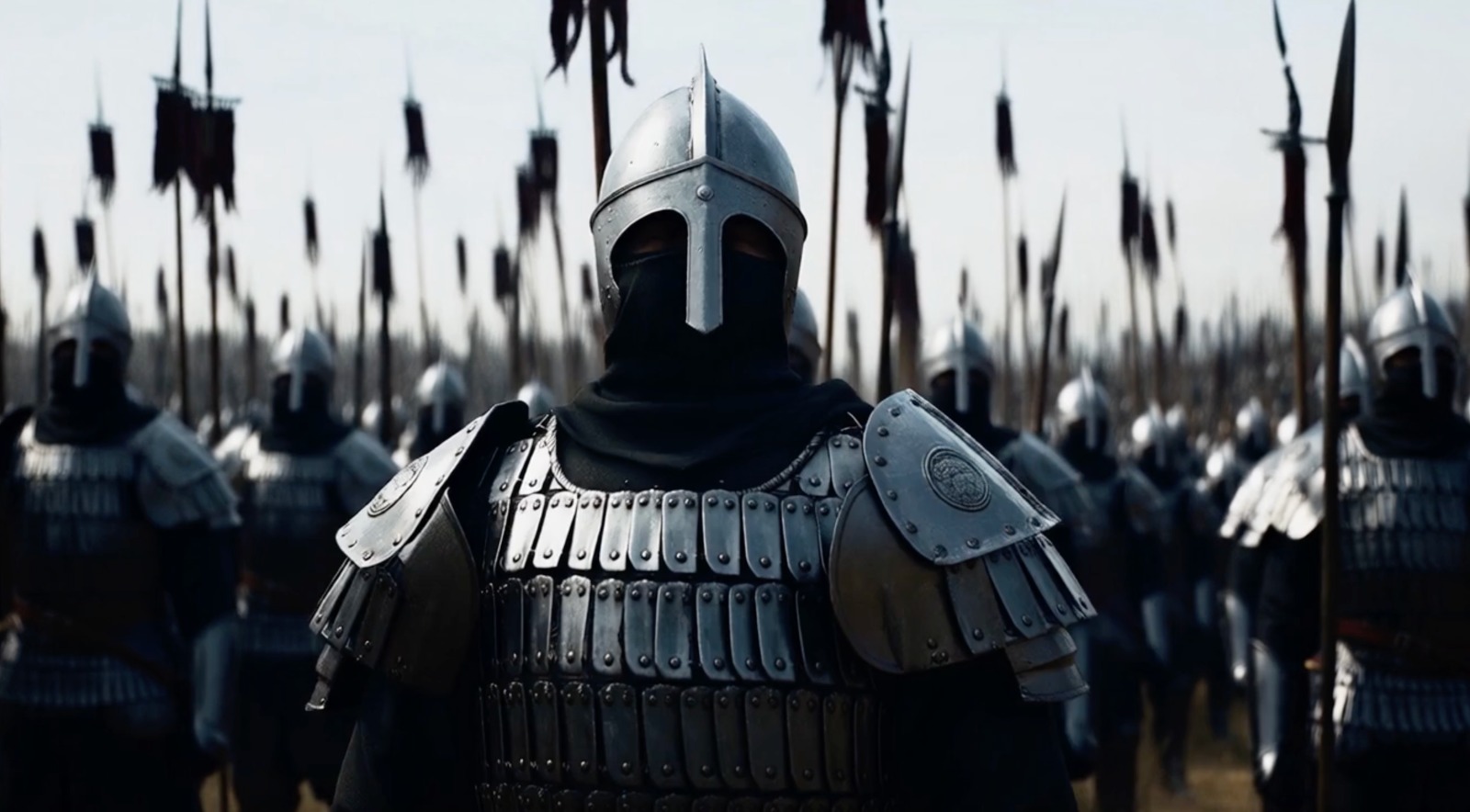}\hfill
\includegraphics[width=0.32\textwidth,height=2.8cm,keepaspectratio]{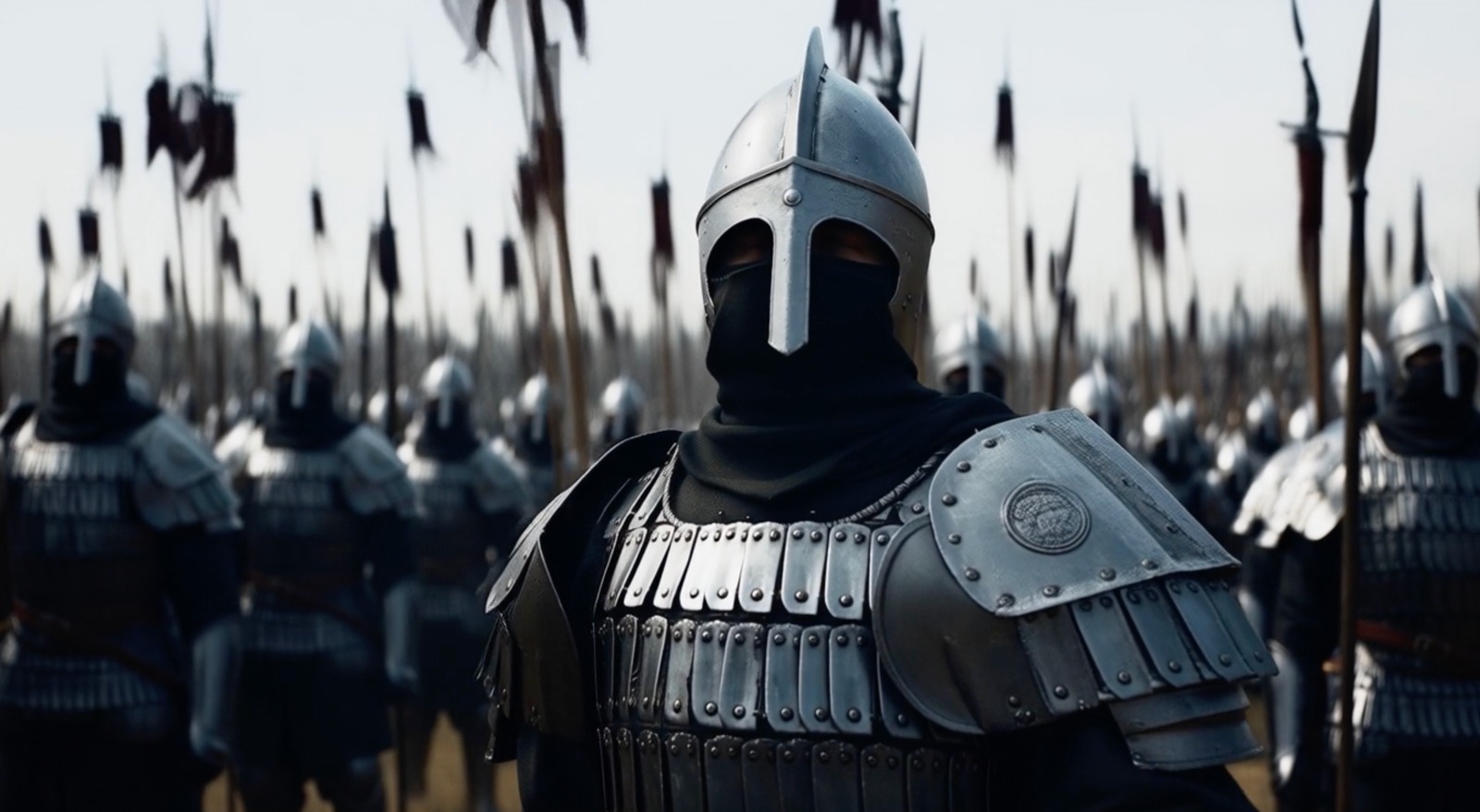}\hfill
\includegraphics[width=0.32\textwidth,height=2.8cm,keepaspectratio]{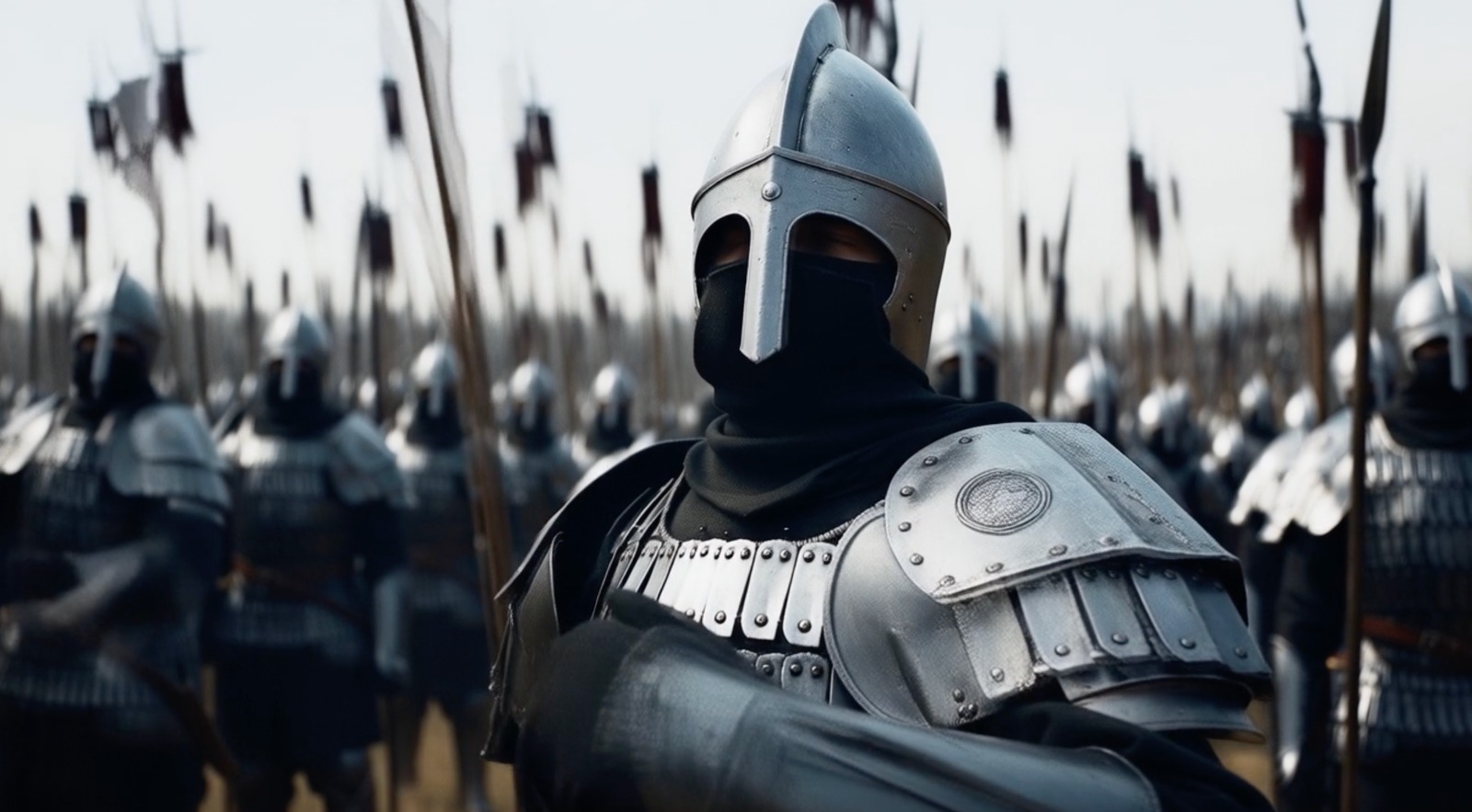}

\vspace{0.15cm}

\includegraphics[width=0.32\textwidth,height=2.8cm,keepaspectratio]{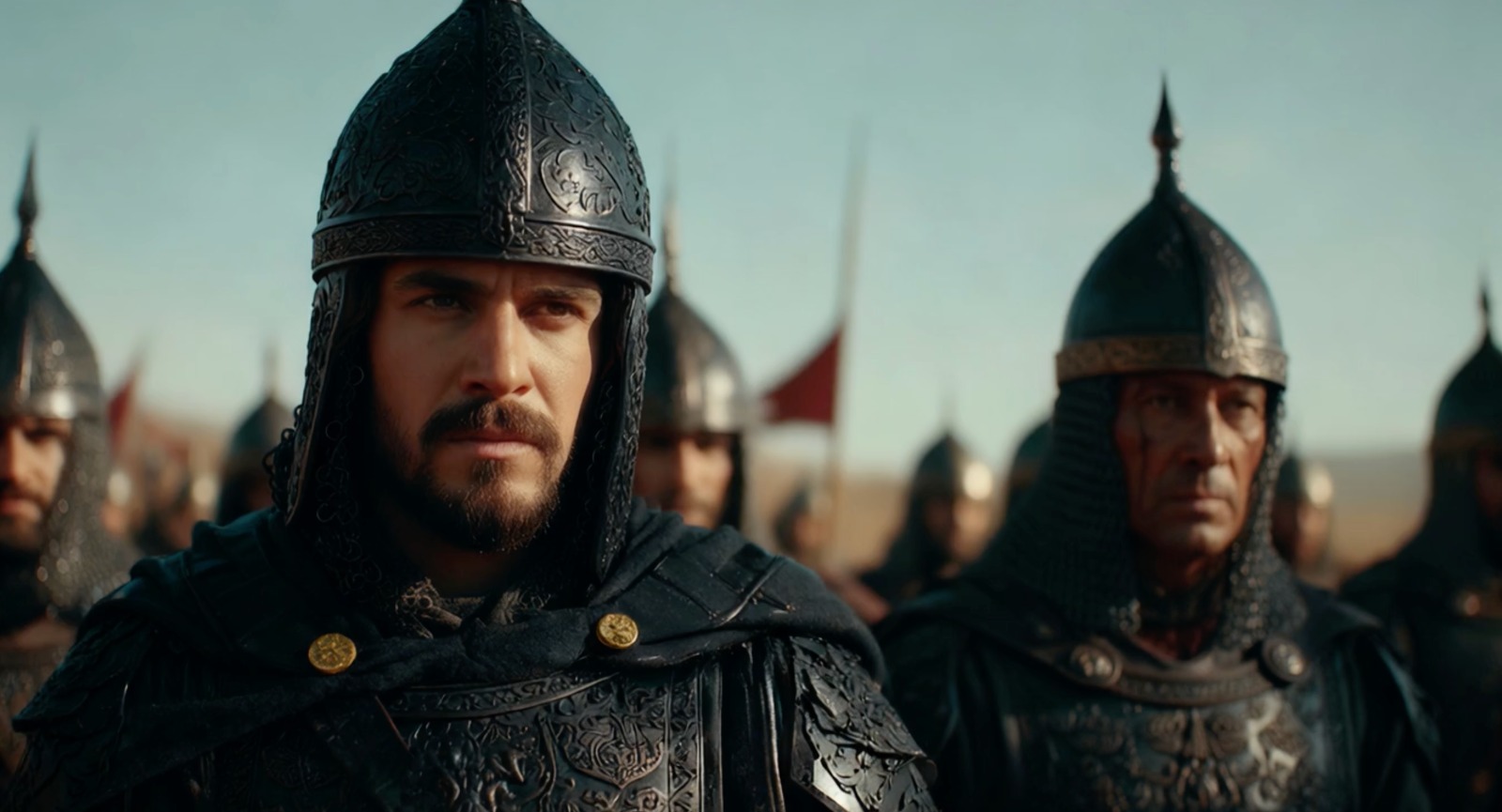}\hfill
\includegraphics[width=0.32\textwidth,height=2.8cm,keepaspectratio]{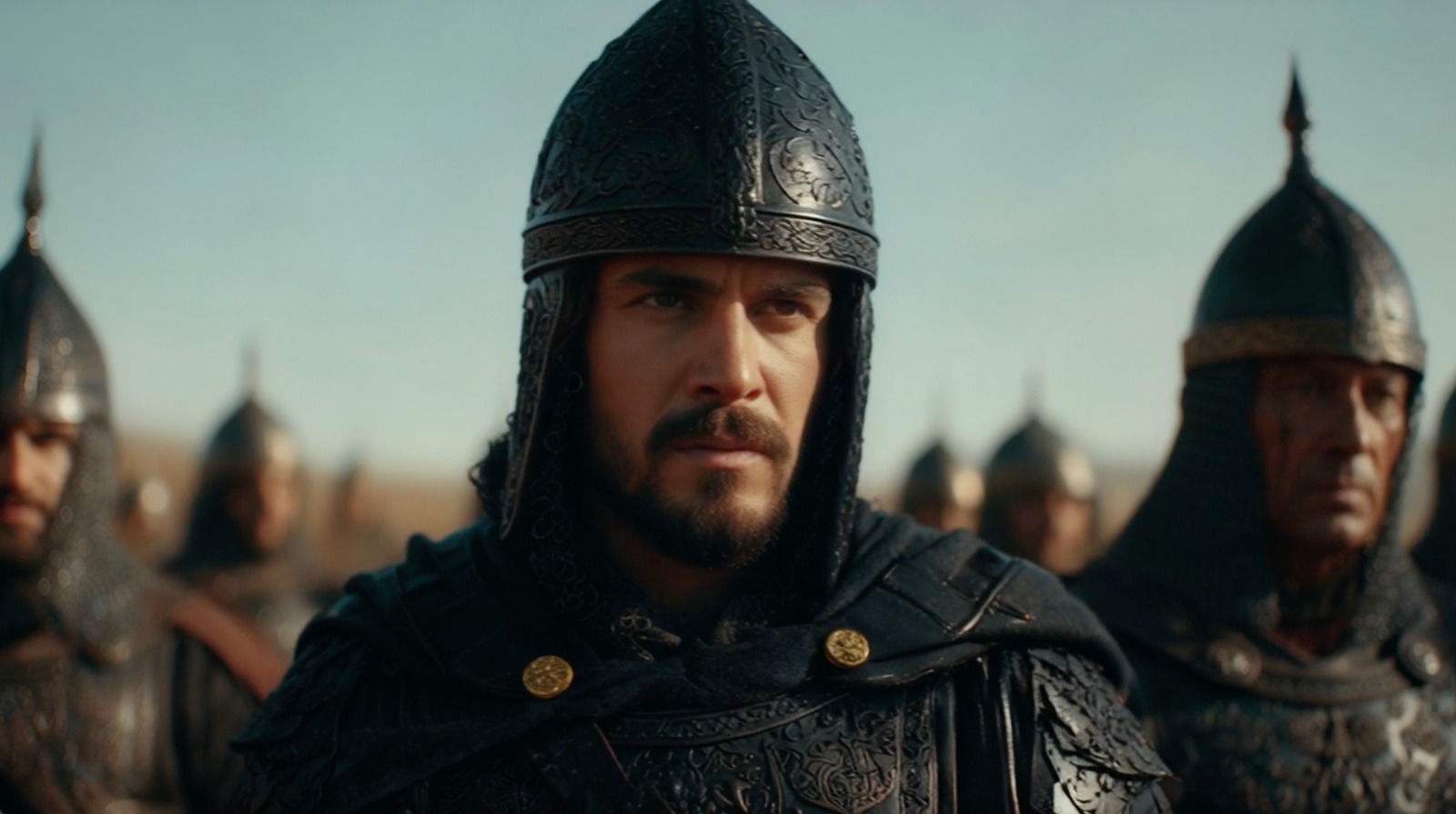}\hfill
\includegraphics[width=0.32\textwidth,height=2.8cm,keepaspectratio]{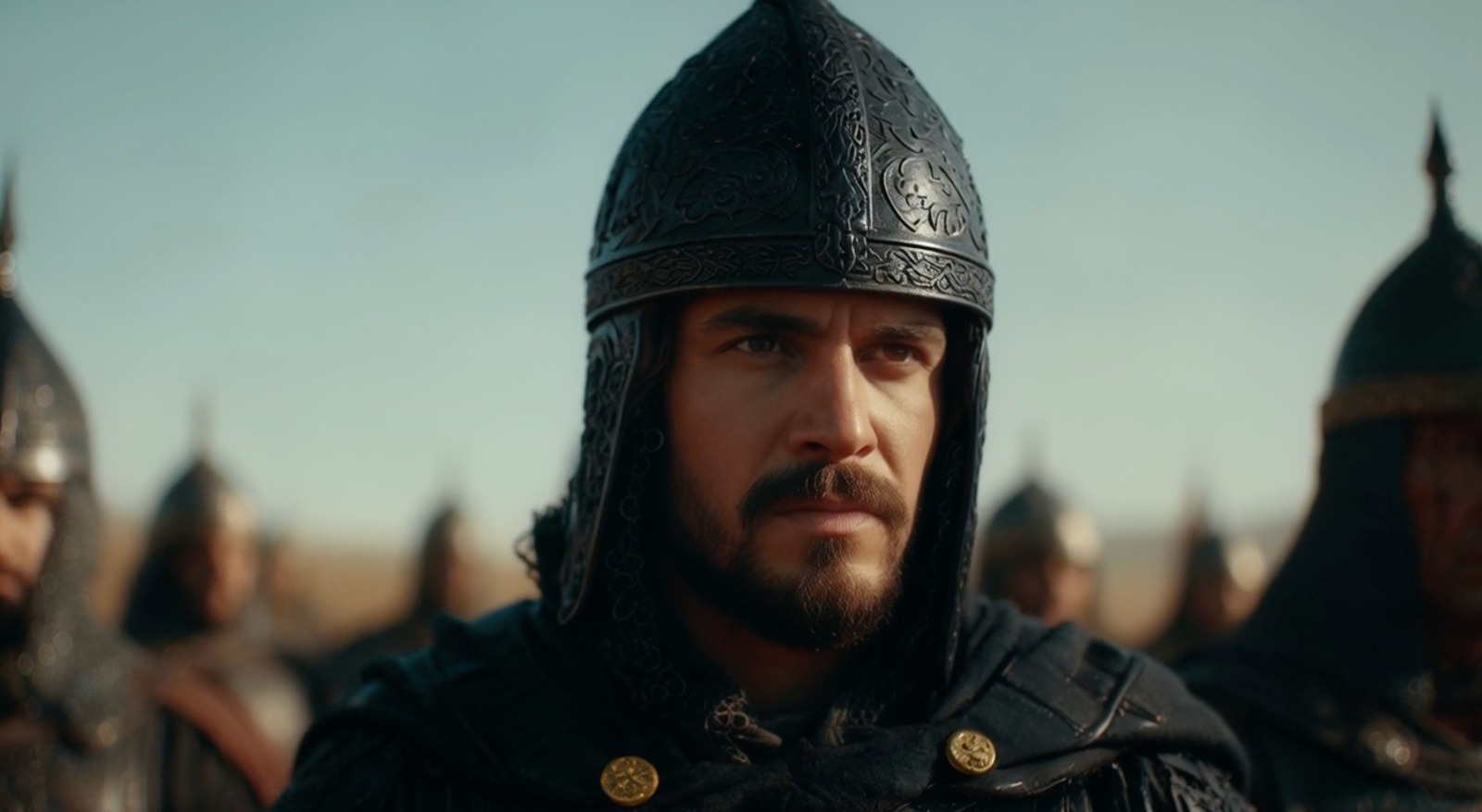}

\vspace{0.15cm}

\includegraphics[width=0.32\textwidth,height=2.8cm,keepaspectratio]{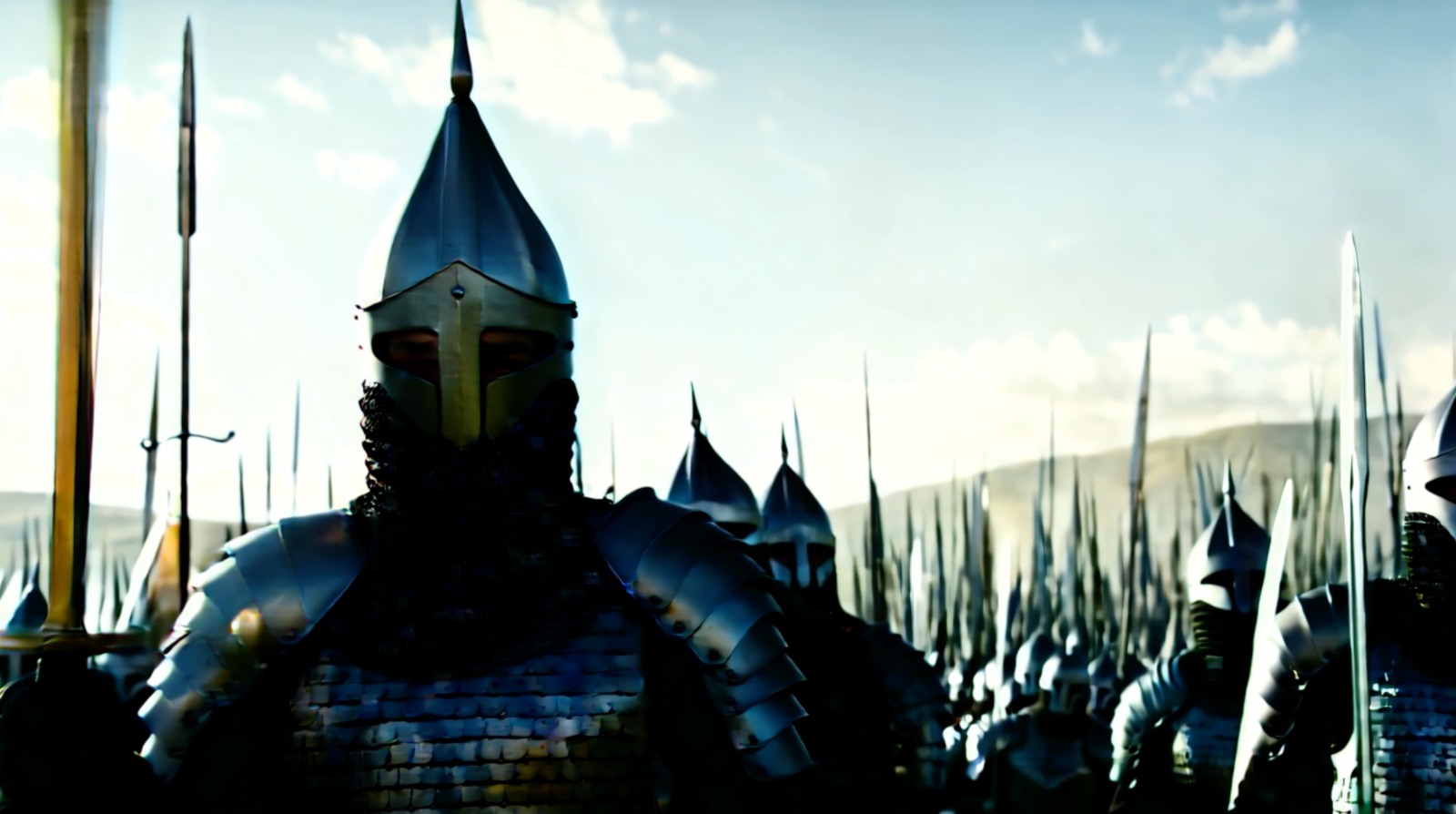}\hfill
\includegraphics[width=0.32\textwidth,height=2.8cm,keepaspectratio]{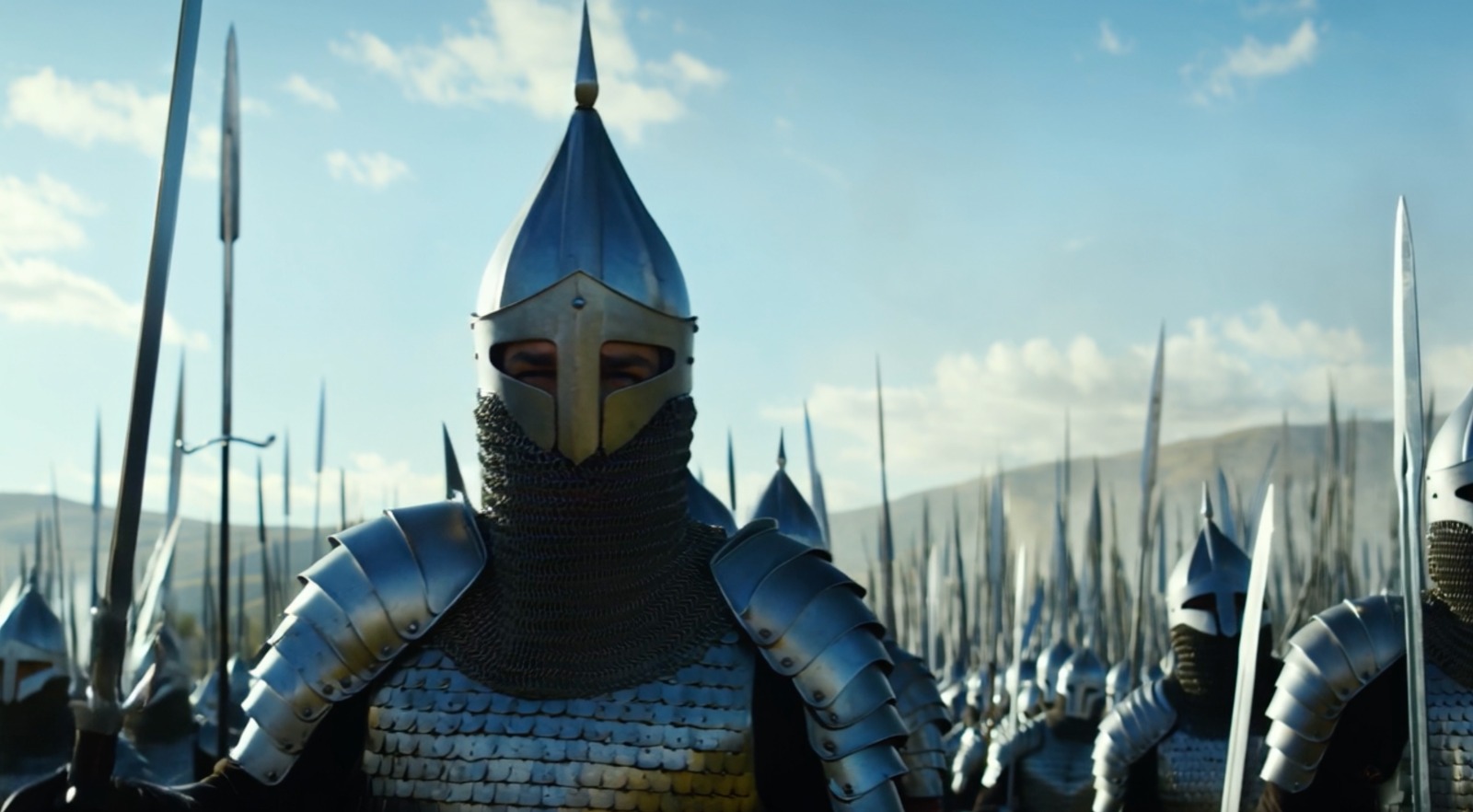}\hfill
\includegraphics[width=0.32\textwidth,height=2.8cm,keepaspectratio]{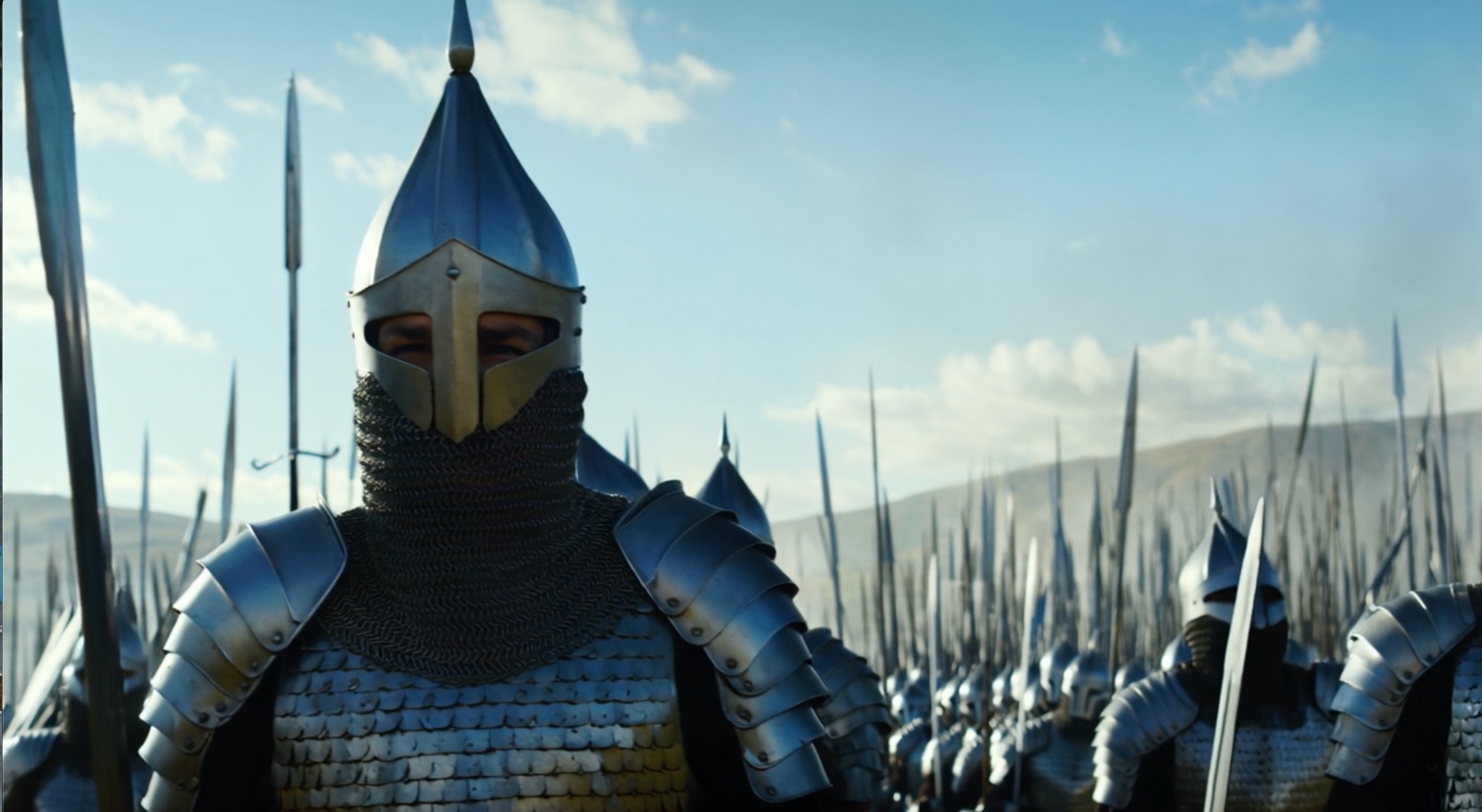}

\vspace{0.15cm}

\includegraphics[width=0.32\textwidth,height=2.8cm,keepaspectratio]{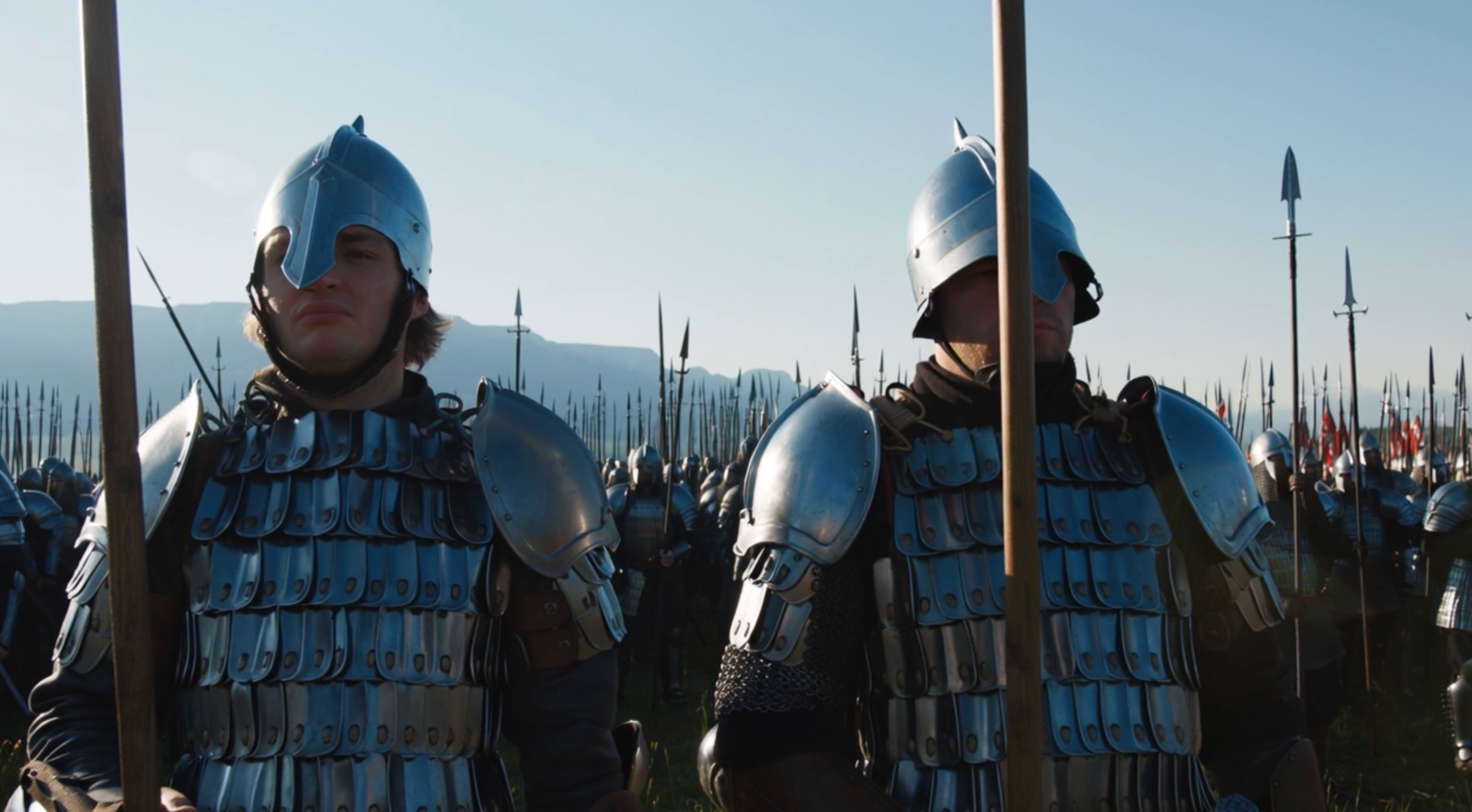}\hfill
\includegraphics[width=0.32\textwidth,height=2.8cm,keepaspectratio]{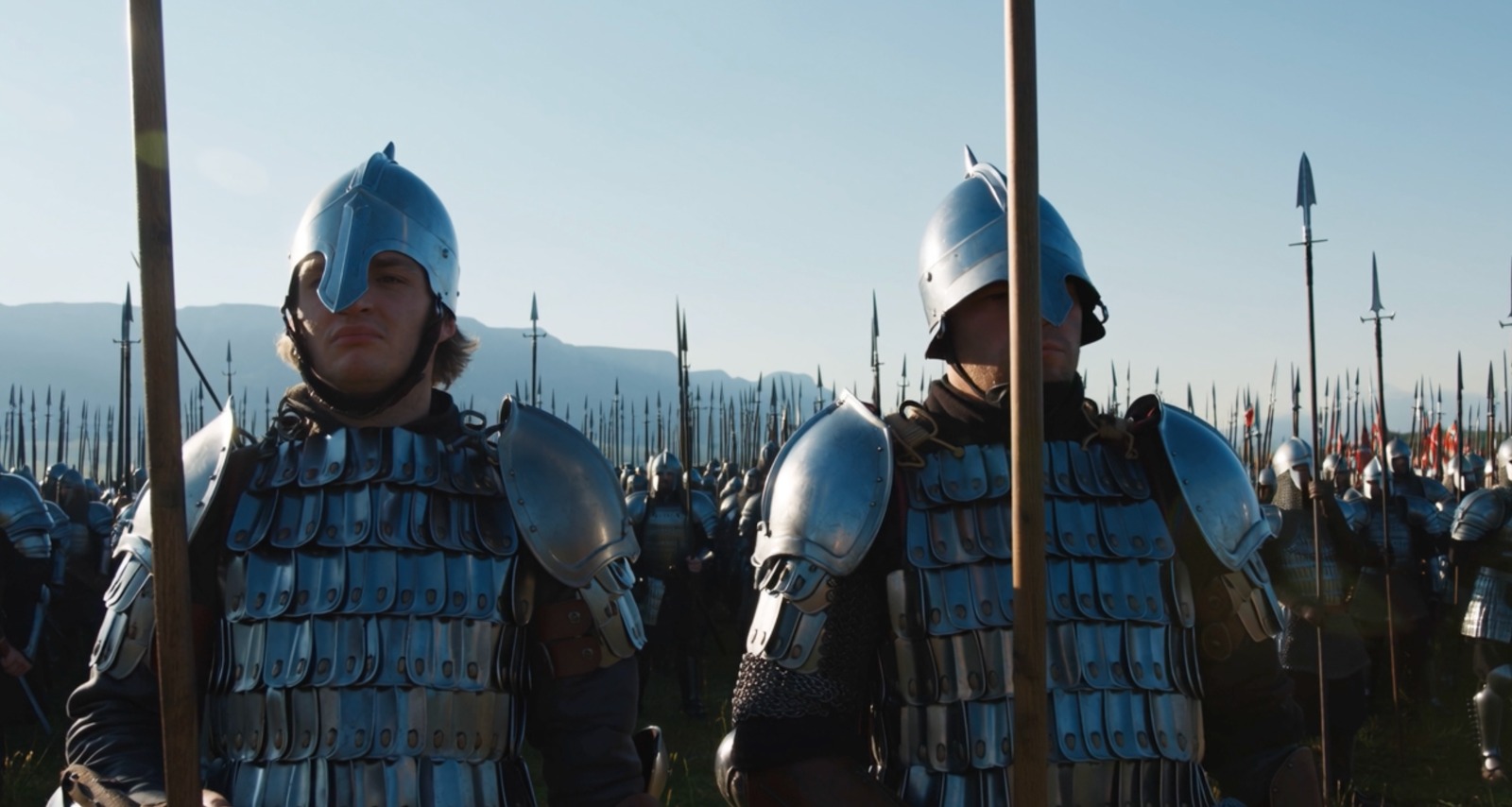}\hfill
\includegraphics[width=0.32\textwidth,height=2.8cm,keepaspectratio]{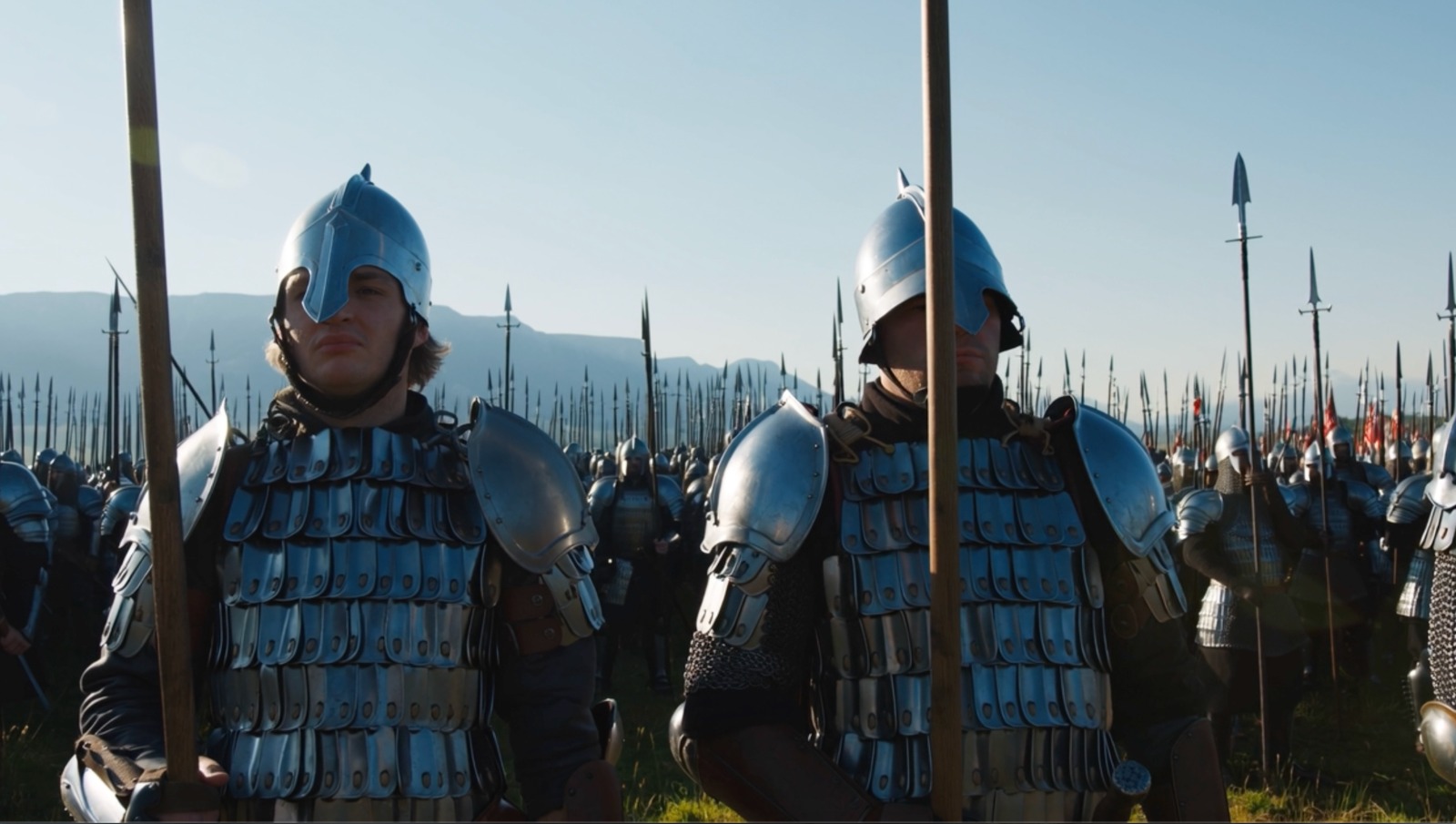}

\vspace{0.15cm}

\includegraphics[width=0.32\textwidth,height=2.8cm,keepaspectratio]{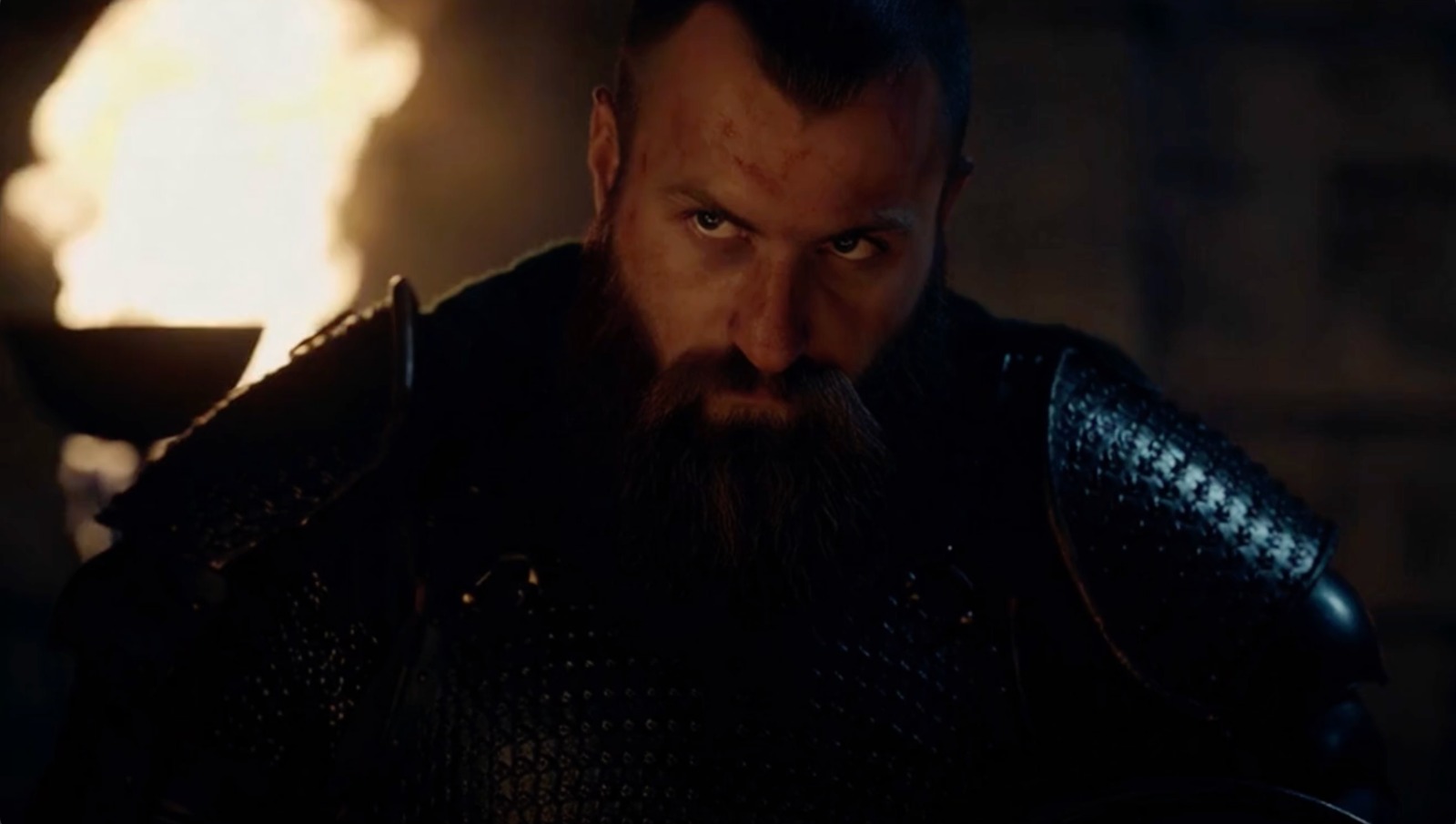}\hfill
\includegraphics[width=0.32\textwidth,height=2.8cm,keepaspectratio]{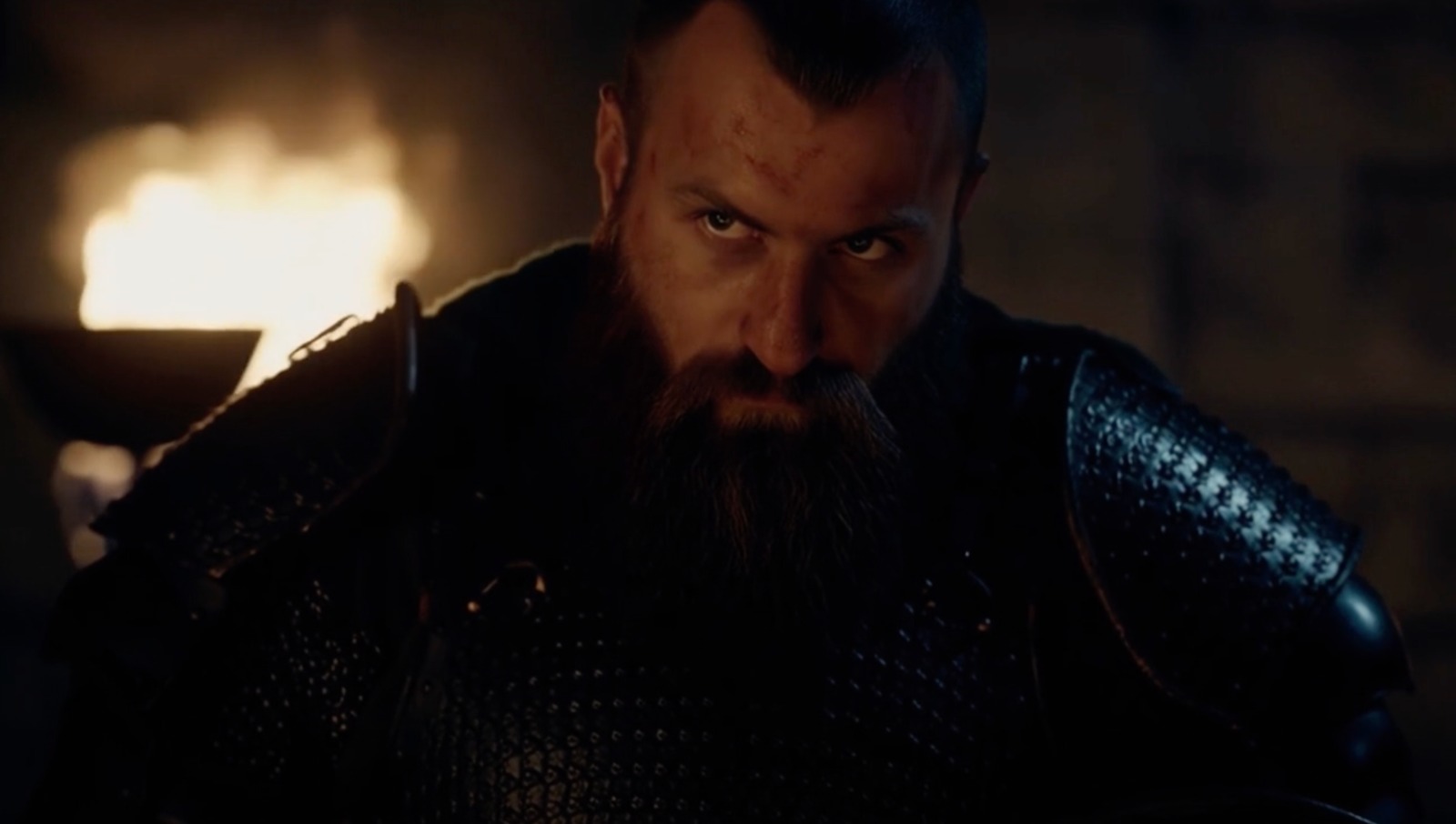}\hfill
\includegraphics[width=0.32\textwidth,height=2.8cm,keepaspectratio]{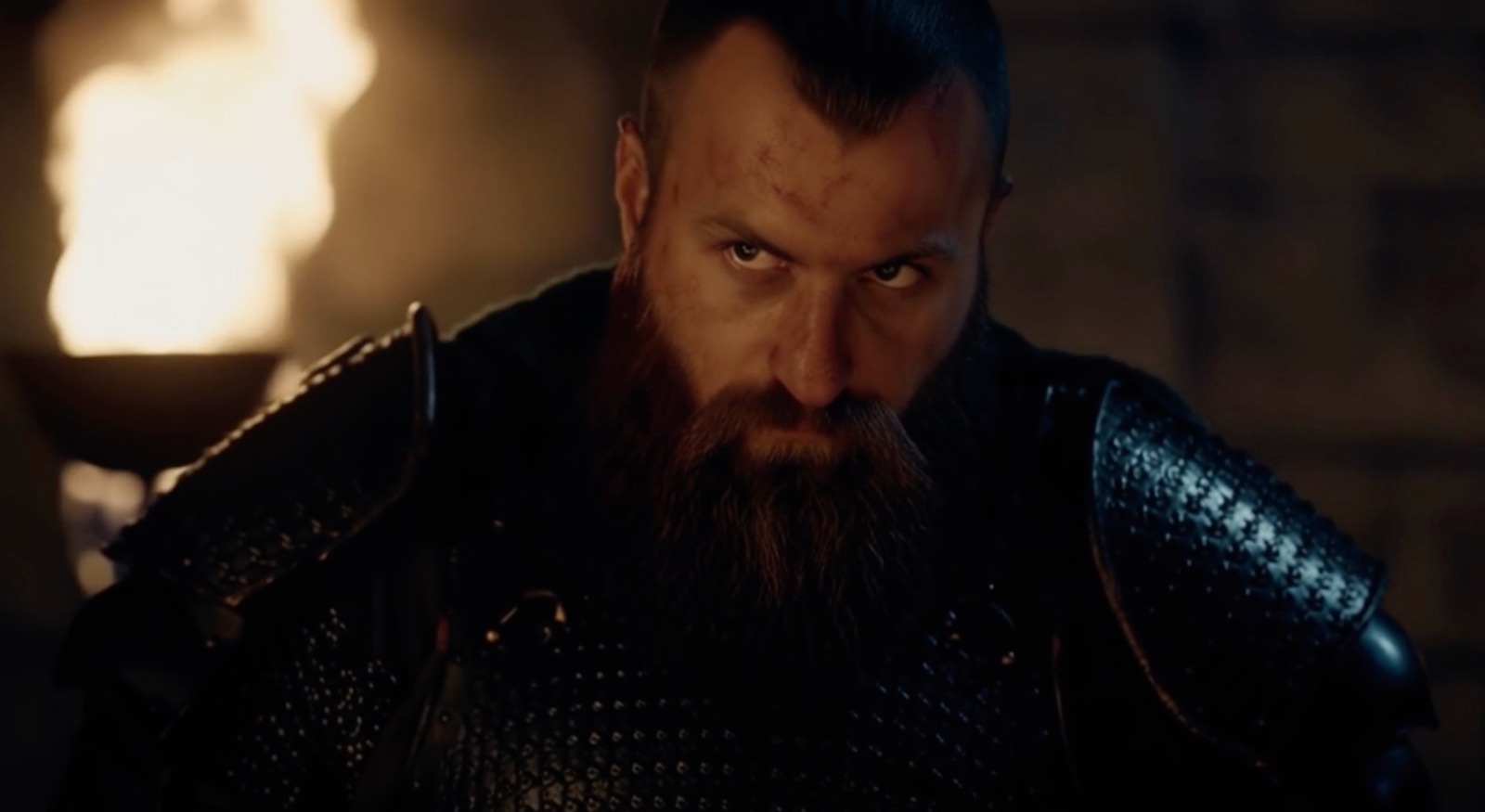}

\vspace{0.15cm}

\caption{\textbf{Comprehensive Visual Results from El Turco Fine-Tuning.} Generated sequences demonstrating temporal coherence and stylistic consistency across diverse scene compositions, camera angles, and lighting conditions. The figure presents 24 frames across 8 sequential rows, illustrating the model's capability to maintain cinematic quality throughout extended sequences. Each row represents a distinct scene or camera angle: close-up helmet details (rows 1--2), wide battlefield formations with atmospheric lighting (rows 3--4), dramatic single-subject shots (rows 5--6), and ensemble compositions with historical armor detail (rows 7--8). All sequences generated at 720p (1280$\times$720) with 30 denoising steps and CFG scale 3.8, demonstrating the model's internalization of \textit{El Turco}'s complete visual grammar while maintaining production-quality cinematography and historical authenticity.}
\label{fig:comprehensive_results}
\end{figure*}

\section{Conclusion}

We presented a practical, reproducible pipeline for adapting large-scale video diffusion transformers to cinematic styles using limited data and accessible hardware. Building on Wan 2.1 I2V-14B, a 14-billion-parameter image-to-video diffusion transformer, we introduce parameter-efficient Low-Rank Adaptation (LoRA) modules to internalize stylistic features from short sequences of the historical television film \textit{El Turco}. The fine-tuned model reproduces historically authentic battlefield and palace scenes while modifying less than 1 \% of the base parameters

Training converges in under two hours on dual A100 GPUs, and multi-GPU inference with Fully Sharded Data Parallelism (FSDP) achieves near-linear speed-up while preserving temporal coherence. Qualitative and ablation studies confirm a balanced trade-off between fidelity and efficiency. The complete open-source pipeline, including preprocessing scripts, training configurations, and inference workflows, bridges state-of-the-art video diffusion research with cinematic production—advancing algorithmic storytelling and creative direction through generative AI.

\section{\textbf{Future Work}}

Our pipeline demonstrates effective cinematic adaptation from limited data, yet several directions remain open. This study focuses on a single historical production, \textit{El Turco}, within a narrow aesthetic range. Extending fine-tuning across other genres—such as science fiction or noir—would test the model’s capacity to generalize and interpolate visual styles. The current 33-frame training and 96-frame inference windows restrict output to brief sequences; generating full scenes will require more memory-efficient, long-context mechanisms.

Text prompting alone offers limited directorial control. Adding spatial or storyboard guidance could enable finer manipulation of framing, lighting, and motion, aligning generative models more closely with real cinematography. Further work should also examine data scaling—training with fewer or more clips—and assess the limits of data efficiency through few-shot adaptation.

Comparisons with open and commercial baselines, and the development of perceptual cinematic metrics for continuity and rhythm, would better situate this work within the field. Finally, testing the pipeline in actual production workflows will clarify its creative and economic value, while transparent standards for consent and attribution remain essential as generative tools approach professional filmmaking quality.

\section*{Acknowledgment}

The authors thank the creators and distributors of the \textit{El Turco} television series for making footage publicly available for research purposes. We acknowledge Google Colab Pro and RunPod for providing affordable GPU compute resources (A100-40GB and dual A100-80GB configurations) that enabled training and inference on a limited budget. We are grateful to the open-source community behind Wan 2.1, Stable Diffusion XL, LoRA, and the DeepSpeed framework for their foundational contributions to video generation and parameter-efficient fine-tuning. Special thanks to the Hagia AI Research Collective for supporting this work and fostering collaborative research in generative AI for cinematic applications.

\end{document}